\documentclass[preprint,12pt]{elsarticle}

\usepackage{amssymb}
\usepackage{setspace}
\usepackage{stmaryrd}
\usepackage{amsfonts}
    \usepackage{algorithmicx}
    \usepackage[ruled]{algorithm}
    \usepackage{algpseudocode}
\usepackage{multirow}
\usepackage{hhline}
\usepackage{fixltx2e}
\usepackage{longtable}
\usepackage{amstext}
\usepackage{caption}
\usepackage{graphicx,times,amsmath} % Add all your packages here
\usepackage{caption}
\usepackage{subcaption}
\usepackage{pdflscape}
\usepackage{afterpage}
\usepackage{capt-of}% or use the larger `caption` package

\usepackage{lipsum}% dummy text
\usepackage{geometry}
\journal{Applied Soft Computing}

\begin{document}

\begin{frontmatter}



\title{Diversity Enhancement for Micro-Differential Evolution}


\author[1]{Hojjat Salehinejad}
\author[1]{Shahryar Rahnamayan}
\author[2]{Hamid R. Tizhoosh}
\address[1]{Department of Electrical, Computer, and Software Engineering, University of Ontario Institute of Technology, 2000 Simcoe Street North, Oshawa, ON L1H 7K4, Canada}
\address[2]{Department of Systems Design Engineering, University of Waterloo, 200 University Avenue West, Waterloo, ON
N2L 3G1, Canada}

\begin{abstract}
The differential evolution (DE) algorithm suffers from high computational time due to slow nature of evaluation. In contrast, micro-DE (MDE) algorithms employ a very small population size, which can  converge faster to a reasonable solution. However, these algorithms are vulnerable to a premature convergence as well as to high risk of stagnation. In this paper, MDE algorithm with vectorized random mutation factor (MDEVM) is proposed, which utilizes the small size population benefit while empowers the exploration ability of mutation factor through randomizing it in the decision variable level. The idea is supported by analyzing mutation factor using Monte-Carlo based simulations. To facilitate the usage of MDE algorithms with very-small population sizes, new mutation schemes for population sizes less than four are also proposed. Furthermore, comprehensive comparative simulations and analysis on performance of the MDE algorithms over various mutation schemes, population sizes, problem types (i.e. uni-modal, multi-modal, and composite), problem dimensionalities, and mutation factor ranges are conducted by considering population diversity analysis for stagnation and trapping in local optimum situations. The studies are conducted on 28 benchmark functions provided for the IEEE CEC-2013 competition. Experimental results demonstrate high performance and convergence speed of the proposed MDEVM algorithm.

\end{abstract}


\end{frontmatter}


\section{Introduction}
Evolutionary algorithms are practical tools for solving real-world problems, \cite{f1}. Accuracy enhancement and increasing the convergence speed toward finding the global solution(s) in optimization algorithms have motivated many researchers to develop more efficient evolutionary approaches. The differential evolution (DE) algorithm is one of the state-of-the-art global optimization algorithms, which is popular due to its simplicity and effectiveness. This algorithm works based on a set of individuals, called population, where an optimal size setting is imperative for a good performance \cite{1}. The centroid-based approach is one of the successful approaches for DE algorithm, which works based on the computing the centroid of population \cite{cen1}, \cite{cen2}. Opposition based computing is another approach which has a great potential in performance improvement of DE algorithm \cite{t2}.

Different variants of DE algorithm with large population size often grant more reasonable results than their small population size versions. A large population size supports a higher diversity for the population, which recombination of its diverse members offers a higher exploration ability to the optimizer to find global solution(s) \cite{2}-\cite{4}. The proposed diversity enhancement technique in this paper offers a better exploration of problem landscape. Most of the research works during past decades were focused on developing complex approaches with a large populations size \cite{38}. Utilizing a large population size intrinsically encompasses more function evaluations and as a consequence, and naturally a lower convergence rate \cite{2}. Therefore, using algorithms with a large population size may not be satisfactory for real-time or on-line applications \cite{37}, \cite{39}. 

Using a population size much smaller than the number of decision variables is sometimes more efficient than some of the state-of-the-art DE algorithms with a large population. The term micro-algorithm, denoted by $\mu$-algorithm, refers to population-based algorithms with a small population size \cite{4}. The micro-algorithms have been used in diverse applications, exceptionally due to their lighter hardware requirements and opportunity to operate in embedded systems with a memory saving approach \cite{1}. Employing small population sizes decreases the number of function calls, but unfortunately due to lack of diversity, it also increases the risk of premature convergence as well as stagnation.

The premature convergence problem refers to the situation where the population has converged to a sub-optimal solution of a multimodal objective function \cite{2}. This situation mostly occurs when the population has lost its diversity and cannot jump out of local optima. In this case, the algorithm progresses slower than usual and may stop any further improvement of the evolved candidate solutions \cite{2}, \cite{39}.  

In stagnation scenario, the population diverts from the correct path toward optimality and not converged to any local optima or any other point as the algorithm proceeds. In this case, the population keeps a certain level of diversity through generations.  Even adding new individuals to the population or updating the current individuals may not guide the algorithm toward convergence \cite{2}. A sign of this behaviour is static proceeding of the found candidate solutions, while the individuals are operating on the problem landscape over generations \cite{39}.  

Based on the stagnation and premature convergence characteristics, it seems reducing the population size while raising the diversity of the population is a key point to achieve a faster convergence speed while maintaining a low risk of premature convergence or stagnation \cite{2}, \cite{39}. The DE algorithm is consisted of several manually tuneable control parameters, where different adaptive proposals have been devised to avoid manual adjustments \cite{61}. One way to increase the diversity of population while keeping its convergence toward global solution(s) is  using intelligent adaptive techniques.

Mutation factor $F$ is one of the critical parameters that is usually set by user \cite{Segura_2015}. A simple modification to overcome the stagnation and pre-mature convergence problems is using randomized values for the mutation factor \cite{Das_2005}, \cite{Das_2005a}, \cite{39}. The authors have recently proposed the idea of vectorized random mutation factor in DE algorithm for each decision variable of the problem, called MDEVM \cite{39}. This algorithm has been recently cited as one of the competitive algorithms in the literature \cite{Brown_2015}. In this paper, we provide a comprehensive survey on micro-EAs. The proposed MDEVM method is discussed in deep and supporting Monte-Carlo simulations to analyze mutation factor diversity are presented. For the first time, we propose a mutation scheme that can work for very-small population size (i.e. $N_{P}=\{2, 3, 4\}$) and comparative analysis on variant problem dimensions and mutation schemes for the MDE algorithms are presented. The considered benchmark problems are a set of 28 functions that cover uni-modal, multi-modal, and composite problems from CEC-2013. The studies are continued on variant ranges for mutation factors, population diversities, and variant stopping conditions for the MDE algorithm. 

In the next section, the micro-population based methods are briefly surveyed. Then, a review of the DE algorithm is presented in Section 3. In Section 4, the proposed method is presented, and diversity enhancement 
in MDE using different structures of mutation factor is studied in detail. The simulation results and corresponding analysis are provided in Section 5. Finally, the paper is concluded in Section 6.

\section{Related Works}

Many research works have attempted to introduce efficient micro-algorithms. The research works can be categorized in four main groups which are micro-genetic algorithms (micro-GAs), micro-particle swarm optimization (micro-PSO), MDE, and other population-based approaches. %A summary of such approaches are provided in Table~\ref{summary}. 

\subsection{Micro-Genetic Algorithms} One of the earlier research work in this direction was a genetic algorithm (GA) with five chromosomes \cite{5}. The strategy in this micro-GA is to copy the best found chromosome from the current population to the next generation. This work was tested on low-dimensional problems, which resulted a faster convergence speed compared to the classical GA. The idea of population reinitialization for micro-GA was another early work in the field \cite{31}. In this approach, the best individual of each converged population, after a predefined number of generations, is replaced with a randomly selected individual in the population of the next iteration. The parallel version of micro-GA, called parallel micro-genetic algorithm (PMGA), was reported in \cite{32}; which solves the ramp rate constrained economic dispatch (ED) problems for generating units with non-monotonically and monotonically increasing incremental cost functions. The PMGA is implemented on a thirty-two-processor Beowulf cluster and the reported results demonstrate feasibility of this approach in online applications. 
The micro-algorithms also have been employed in multi-objective optimization (MOO). The improved version of nondominated sorting genetic algorithm (NSGA-II) with a specific population initialization strategy are embedded into the standard micro-GA to solve the MOO problems \cite{10}. A micro-GA with a population size of four and a reinitialization strategy is proposed in \cite{28} which can produce a major part of the Pareto front at a very low computational cost. Three forms of elitism and a memory are used to generate the initial population \cite{28}. An improved version of micro-GA, archive-based micro-GA (AMGA2) for constrained MOO is proposed in \cite{33}. This algorithm is based on a steady-state GA that preserves an external archive of best and divert candidate solutions. This small population-based approach facilitates the decoupling of the working population, the external archive, and the number of required solutions as the outcome of the optimization procedure. A model of MOO for hierarchical GA (MOHGA) based on the micro-GA approach for modular neural networks (MNNs) optimization is proposed in \cite{34}. This approach is used in iris recognition. The MOHGA divides the input data into granules and sub-modules and then decides to split the data for training and testing phases. It is reported that this technique can obtain good results based on using less data \cite{34}. The micro-GA has also been used for local fine tuning in an adaptive local search intensity manner for training recurrent artificial neural networks (ANN) \cite{35}. It is reported that this approach is useful for systems identification tasks. In \cite{21} a multi-objective micro genetic extreme learning machine (MOMG-ELM) is proposed, which provides the appropriate number of hidden nodes in the machine for solving the problem, which minimizes the mean square error (MSE) of the training phase. The micro-GA is applied successfully for many applications such as designing wave-guide slot antenna with dielectric lenses \cite{36}, detection of flaws in composites \cite{30}, and scheduling of a real-world pipeline network \cite{29}, where better performances compared to the standard GA are reported.

\subsection{Micro-Particle Swarm Optimization} The particle swarm optimization (PSO) is one of the well-known swarm intelligence algorithms, which its small population size versions have been developed \cite{6}, \cite{51}, \cite{7}. The Coulomb's law is used in micro-PSO method for high dimensional problems \cite{6} . First achievement of this approach is removal of the burden for determining the suitable size of space needed to enclose the blacklisted solutions and the amount of repulsion needed to repel the particles, as these parameters are extremely difficult to determine for high dimensional problems. The other achievement is the flexibility of controlling the repulsion on particles through the use of a parameter which controls the amount of repulsion experienced by the particles at a particular position.The conducted simulations' results on five high-dimensional benchmark functions demonstrate superior performance of micro-PSO versus the standard PSO with a large populations size. A five-particle micro-PSO is used in \cite{41} to deal with constrained optimization problems. This method preserves population diversity by using a reinitialization process and incorporates a mutation operator to improve the exploratory capabilities of the algorithm. The reported results present competitive performance versus the simple multi-membered evolution strategy (SMES) and stochastic ranking (SR) method \cite{41}. The micro-PSO was employed for MOO in \cite{42}; it produces reasonably good approximations of the Pareto front of moderate dimensional problems with a small number of objective function evaluations (only 3000 calls per run), comparing to PSO approach.
In another micro-PSO algorithm, a parallel master-slave model of cooperative micro-PSO was introduced \cite{7}, in which the original search space is decomposed into subspaces with smaller dimensions. Then, five individuals are considered in each subspace to identify suboptimal partial solution components. Its performance was assessed on a set of five widely used test problems with significant improvements in solution quality, compared to the standard PSO algorithm \cite{7}. A cooperative PSO approach was proposed in \cite{47} which uses a company of low-dimensional and low-cardinality sub-swarms to deal with complex high-dimensional problems. Promising results are reported using this methods, tested with five widely used test problems. A clonal selection algorithm (CSA), which belongs to the family of artificial immune system (AIS), in conjunction with a micro-PSO (CS$^{2}$P$^{2}$SO) was introduced in \cite{46} as a hybrid scheme. In this hybridization, the strength of standard PSO algorithm is enhanced, where the micro-PSO helps to find the optimum solution with less memory requirement and the CSA increases the exploration capability while reducing the chance of convergence to a local minima. Simulations are conducted on only four benchmark functions, where competitive performance is reported. A mixed-integer-binary small-population PSO was proposed in \cite{58} for solving a problem of optimal power flow. The constraint handling technique used in this algorithm is based on a strategy to generate and keep its four decision variables in feasible space through heuristic operators. In this way, the algorithm focuses its search procedure on the feasible solution space to obtain a better objective value. This technique improves the final solution quality as well as the convergence speed \cite{58}. The micro-PSO has been developed for many applications such as motion estimation \cite{40}, power system stabilizers design \cite{43}, \cite{45}, optimal design of static var compensator (SVC)  damping controllers \cite{44}, reactive power optimization \cite{48}, short-term hydrothermal scheduling \cite{50}, reconfiguration of shipboard power system \cite{52}, and transient stability constrained optimal power flow \cite{49}.
 
\subsection{Micro-Differential Evolution Algorithms}
The DE algorithm works based on the scaled difference between two individuals of a population set, where the scaling factor is called the mutation factor. Due to reliability and simplicity of the DE algorithm, it has been employed in many science and engineering areas such as, solving large capacitor placement problem \cite{17} and synthesis of spaced antenna arrays \cite{18}. Many works have put new schemes forward to enhance the DE algorithm such as, opposition-based differential evolution (ODE) \cite{14}, enhanced differential evolution using center-based sampling \cite{15}, and opposition-based adaptive differential evolution \cite{16}. Some approaches toward reducing computational cost of DE-based algorithms by reducing the population size have been proposed as well, \cite{8}-\cite{9}, \cite{11}-\cite{13}. In order to increase the exploration ability of MDE algorithm and to prevent stagnation, an extra search move is incorporated into the MDE algorithm in \cite{1} by perturbing it along the axes. A local search procedure is hybridized with the MDE algorithm in \cite{3} to overcome high dimensional problems. However, the reported performance results are comparable with some other methods. As an application of MDE, a hybrid differential evolution (HDE) with population size of five is used for finding a global solution \cite{60}. A gradually reducing population size method is proposed in \cite{8}. This method is examined on 13 benchmark functions, where the results have demonstrated a higher robustness as well as efficiency compared to the parent DE \cite{8}. In another approach \cite{9}, small size cooperative sub-populations are employed to find sub-components of the original problem concurrently. During cooperation of sub-populations, the sub-components are combined to construct the complete solution of the problem. Performance evaluation of this method has been done on high-dimensional instances of five sample test problems with encouraging results reported in \cite{9}. MDE is employed for evolving an indirect representation of the bin packing problem with acceptable performance \cite{11}. The idea of self-adaptive population size was carried out to test absolute and relative encoding methods for DE \cite{12}. The reported simulation results on 20 benchmark problems denote that, in terms of the average performance and stability, the self-adaptive population size using relative encoding outperforms the absolute encoding method and the standard DE algorithm \cite{12}. The idea of micro-ODE was proposed and evaluated for an image thresholding case study \cite{13}. Performance of the proposed method was compared with the Kittler algorithm and the MDE. The micro-ODE method has outperformed these algorithms on 16 challenging test images and has demonstrated faster convergence speed due to embedding the opposition-based population initialization scheme \cite{13}. The smallest population size used in \cite{Fajfar_2012} is $N_{P}=10$. This method tries to decrease the population size by using three different rules to select candidates for replacing the trial vector \cite{Fajfar_2012}. 

It is worth mentioning that MDE is different from the compact DE (cDE) methods \cite{Mininno_2011}. In cDE methods, a statistical representation of population is used, where the memory requirement is similar to using four individuals in the population, regardless of the problem's dimension \cite{Mininno_2011}, \cite{Brown_2015}. Since in this work we are focused on discussing small non-virtual populations, this class of DE algorithms is left for further investigation in other works.

Many methods have been proposed in the literature to increase robustness and reliability of DE algorithm through adaptive or self-adaptive approaches \cite{Neri_2010}, \cite{Neri_2010a}, \cite{Mininno_2011}. This is particularly important for the hyper-parameters adjustment. The mutation factor is one of those parameters which generally is set to a constant value \cite{Segura_2015}. However, it has been shown that randomization of mutation factor can offer a potential new search moves and compensate the excessively deterministic search structure of a standard DE algorithm \cite{Das_2005}, \cite{39}. Studies have used various distribution such as Gaussian, Log-normal, and Cauchy to generate random mutation factor. However, none of them is superior over the others \cite{Segura_2015}. 

The methods proposed in \cite{Das_2005}, \cite{Das_2005a} use random mutation factor at each generation to increase diversity of the population, which reportedly is effective for both noise and stationary problems \cite{Price_2005}. These methods use a standard population size whereas the mutation factor $F$ is randomly selected from the range $(0.5,1)$ such that its mean value is controlled to remain at 0.75. In \cite{Weber_2011}, four different mutation factor (scale-factor) schemes are proposed. The population size is set to $N_{P}=200$. The study shows that none of the methods can show promising results for all problems, since the performance is dependent on the employed type of the distribution \cite{Weber_2011}. In \cite{Brest_2008}, a self-adapted DE algorithm for the mutation factor and crossover rate parameters is presented. The smallest population size used in the experiments is $N_{P}=25$. A self-adaptive control mechanism is used in \cite{Brest_2009} to change the mutation factor $F$ and crossover rate $C_{r}$ during the generations. In this method, only the ``rand/1/bin'' mutation vector is used for a multi-population method with aging mechanism. The jDE method is one of the promising methods, as $F$ is generated with a specific ratio for each individual of a standard population size \cite{Brest_2006}. The idea of generating random mutation factor at the lowest level (for each individual of population and dimension of problem per generation) is proposed by authors in \cite{39}. This technique is used to increase search performance of the standard MDE algorithms. This method is evaluated on a set of 28 benchmark functions for CEC-2013 competition, where the results show superior exploration performance. This algorithm, called MDEVM, is used as a measure to compare the performance of a new mutation factor, called current-by-rand-to-pbest, proposed in the $\mu$JADE algorithm \cite{Brown_2015}. In this approach, which is a DE algorithm for unconstrained optimization problems, the smallest considered population size is $N_{P}=8$ \cite{Brown_2015}. In this method, the mutation factor $F$ and crossover rate $C_{r}$ are randomly generated at the \textit{beginning of each generation}, where the mean of distribution is updated in each generation. The proposed mutation factor, called current-by-rand-to-pbest, in \cite{Brown_2015} is tested for both large and small population sizes on a set of 13 classical benchmark functions. The comparative results in \cite{39} show competitive performance between MDEVM and $\mu$JADE algorithms. The MDEVM algorithm is developed for 3-D localization of wireless sensors for real-world application in \cite{PIMRC}.

\subsection{Other Micro-Population-based Algorithms}
Several other types of micro-population-based algorithms have been proposed in the literature. A cooperative micro-artificial bee colony (CMABC) approach for large-scale optimization was presented in \cite{53}. This approach has combined the divide-and-conquer property of cooperative algorithms and low computational cost of micro-artificial bee colony (MABC) method. In case of employing micro-bacterial foraging optimization algorithms ($\mu$-BFOA) for solving optimization problems, in \cite{54} the best bacterium is kept unaltered, whereas the other population members are reinitialized. It is reported that this approach has outperformed the standard bacterial foraging optimization algorithm (BFOA) with a larger population size \cite{54}. For the environmental economic dispatch case study, a chaotic micro bacterial foraging algorithm (CMBFA) with a time-varying chemotactic step size is proposed in \cite{55}. It is reported that the convergence characteristic, speed, and solution quality of this method are better than the classical BFOA for a 3-unit system and the standard IEEE 30-bus test system. A micro-artificial immune system (Micro-AIS) with five individuals (antibodies), from which only 15 clones are obtained is proposed in \cite{56}. In this approach, the diversity is preserved by considering two simple but fast mutation operators in a nominal convergence manner, that work together in a reinitialization process \cite{56}. An other type of EAs, called elitistic evolution (EEv), is proposed for optimizing high-dimensional problems in \cite{57}, which works without using complex mechanisms such as Hessian or covariance matrix. This approach utilizes adaptive and elitism behaviour, in which a single adaptive parameter controls the evolutionary operators to provide reasonable local and global search abilities \cite{57}. An efficient scheduler for heterogeneous computing (HC) and grid environments, based on parallel micro-cross generational elitist selection, heterogeneous recombination, and cataclysmic mutation, called p$\mu$-CHC is proposed in \cite{59}. This method combines a parallel sub-populations model with a focused evolutionary search using a micro population and a randomized local search (LS) method. Performance comparisons of algorithms such as ant colony algorithm (ACO) and GA have demonstrated good scheduling in reduced execution times \cite{59}.

%
%\begin{table}[t]
%\caption{Summary of related works in micro-population-based algorithms.}
%\begin{center}
%\footnotesize
%\begin{tabular}{|c|c|c|}
%\hline
%\multicolumn{1}{|c|}{Population-based Algorithm}% time}
%& \multicolumn{1}{|c|}{Related Research Works} \\% cost}\\
% \hline
% \hline
%Genetic Algorithm (GA) & \cite{5},\cite{10}, \cite{21}, \cite{28}-\cite{36}\\ \hline
%Particle Swarm Optimization (PSO) & \cite{6}, \cite{7}, \cite{40}-\cite{52}, \cite{58} \\ \hline
%Differential Evolution (DE) & \cite{1}, \cite{3},\cite{8},\cite{9},\cite{11}-\cite{18},\cite{Fajfar_2012},\cite{Brown_2015}, \cite{39}, \cite{60} \\ \hline
%Artificial Bee Colony (ABC) & \cite{53}\\ \hline
%Bacterial Foraging Optimization (BFO) & \cite{54}\\ \hline
%Artificial Immune System (AIS) & \cite{56}\\ \hline
%Elitistic Evolution (EEv) & \cite{57}\\ \hline
%\end{tabular}
%\label{summary}
%\end{center}
%\end{table}

\section{Differential Evolution}

Generally speaking, while solving a black-box problem to find optimal decision variables, an optimizer has no knowledge about the structure of the problem landscape to minimize/maximize an objective function. The DE algorithm, similar to other algorithms in its category, starts its search procedure with some uniform random initial vectors and tries to improve them in each generation toward an optimal solution. The population $\textbf{\textbf{P}}=\{\textbf{X}_{1},...,\textbf{X}_{N_{P}}\}$ consists of $N_{P}$ vectors in generation $g$, where $\textbf{X}_{i}$ is a $D$-dimensional vector defined as $\textbf{X}_{i}=(x_{i,1},...,x_{i,D})$. Generally a simple DE algorithm consists of the following three major operations: mutation, crossover, and selection.

\textit{Mutation:} This step selects three vectors randomly from the population such that $i_{1} \neq i_{2}\neq i_{3} \neq i$ where $i\in\{1,...,N_{P}\}$ and $N_{P}\ge 4$, for each vector $\textbf{X}_{i}$, the mutant vector scheme ``DE/Rand/1" is calculated as
\begin{equation}
\textbf{V}_{i}=\textbf{X}_{i_{1}}+F(\textbf{X}_{i_{2}}-\textbf{X}_{i_{3}}),
\label{eq:de/rand/1}
\end{equation}  
where the factor $F\in(0,2]$ is a real constant number, which controls the amplification of the added differential vector of $(\textbf{X}_{i_{2}}-\textbf{X}_{i_{3}})$. The exploration ability of DE increases by selecting higher values for $F$. So far, four main mutation schemes are introduced \cite{61},\cite{62}, summarized as

\begin{itemize}
\item DE/Best/1:
\begin{equation}
\textbf{V}_{i}=\textbf{X}_{{best}}+F(\textbf{X}_{i_{1}}-\textbf{X}_{i_{2}})
\end{equation}
\item DE/Target-to-Best/1 (DE/T2B/1):
\begin{equation}
\textbf{V}_{i}=\textbf{X}_{i}+F(\textbf{X}_{{best}}-\textbf{X}_{i})+F(\textbf{X}_{i_{1}}-\textbf{X}_{i_{2}})
\end{equation}
\item DE/Rand/2:
\begin{equation}
\textbf{V}_{i}=\textbf{X}_{i_{1}}+F(\textbf{X}_{i_{2}}-\textbf{X}_{i_{3}})+F(\textbf{X}_{i_{4}}-\textbf{X}_{i_{5}})
\end{equation}
\item DE/Best/2:
\begin{equation}
\textbf{V}_{i}=\textbf{X}_{{best}}+F(\textbf{X}_{i_{1}}-\textbf{X}_{i_{2}})+F(\textbf{X}_{i_{3}}-\textbf{X}_{i_{4}}),
\end{equation}
\end{itemize}
where $\textbf{X}_{best}$ is corresponding vector of the best objective value in the population.

\textit{Crossover:} The crossover operation increases diversity of the population by shuffling the mutant and parent vector as follows:
\begin{equation}
U_{i,d}= \left\{ \begin{array}{ll} V_{i,d}, & rand_{d}(0,1)\le C_{r}\: or\: d_{rand}=d\\
x_{i,d}, & otherwise
\end{array},\right.
\end{equation}
where $d=1,...,D$, is the dimension and $C_{r}\in[0,1]$ is the crossover rate parameter, and $rand(a,b)$ generates a real random uniform number in the interval $[a,b]$. Therefore, the trial vector $\textbf{U}_{i} \; \forall i \in\{1,...,N_{P}\}$ can be generated as
\begin{equation}
\textbf{U}_{i}=(U_{i,1},...,U_{i,D}).
\end{equation}

\textit{Selection:} The $\textbf{U}_{i}$ and $\textbf{X}_{i}$ vectors are evaluated and compared with respect to their fitness values; the one with better fitness value is selected for the next generation.

\section{Proposed Diversity Enhancement via Vectorized Random Mutation}

%\subsection{Proposed Micro-Differential Evolution with Vectorized Random Mutation Factor}

In our proposed algorithm, the population size is very small compared to the standard DE algorithm. Reducing the population size results a faster convergence rate with a higher risk of stagnation. However, by increasing the population diversity it is possible to decrease the stagnation risk \cite{2}, \cite{3}. 
In order to foster diversity, the mutation factor $F$, as one of the most significant control parameters for the DE algorithm, can play a major role. The mutation factor $F$ in the DE algorithm is a constant mutation factor (CMF) generally set to $F=0.5$ \cite{2}, \cite{14}. This factor can also be selected randomly from the interval $[0,2]$ for each individual $i$ in the population vector, $F_{i}=rand(0,2)$, \cite{3}. Different versions of this scalar random mutation factor (SRMF) is proposed in literature for standard DE algorithm as discussed in the previous subsection. We call its micro version as the micro-differential evolution with scalar random mutation factor (MDESM) where the population size is very small as well. In the MDE algorithm, in order to increase the population diversity, we propose the idea of utilizing a vectorized random mutation vector (VRMF) for each individual in the population. This approach is called the MDEVM algorithm. Therefore, the mutation factor can be defined for each individual $i$ as

\begin{equation}
\textbf{F}_{i}=\{F_{i,1},...,F_{i,D}\},\: \forall i \in \{1,...,N_{P}\},
\label{Proposed_F}
\end{equation}  
where $F_{i,j}=rand(0.1,1.5),\: \forall j \in \{1,...,D\}$, \cite{3}. This interval is selected based on the experimenal results presented in the next section. 

\alglanguage{pseudocode}
\begin{algorithm}[!htp]
\small
\caption{\textbf{Micro-Differential Evolution with Vectorized Mutation (MDEVM)}}
\begin{algorithmic}[1]
\State \textbf{Procedure} MDEVM
\State $g=0$

\textit{//\textbf{Initial Population Generation}}

\For {$i=1 \to N_{P}$}
\For {$d=1 \to D$}
\State $\textbf{X}_{i,d}=x_{d}^{min}+rand(0,1) \times (x_{d}^{max}-x_{d}^{min})$
\EndFor
\State $\textbf{P}_{i}^{g}=\textbf{X}_{i}$
\EndFor

\textit{//\textbf{End of Initial Population Generation}}

\While {($|BFV-VTR|>EVTR$ \& $NFC<NFC_{Max}$)}
\For {$i=1 \to N_{P}$}

\textit{//\textbf{Mutation}}

     \State \textit{Select three random population vectors from} $\textbf{P}^{g}$ \textit{where} $(i_{1} \neq i_{2}\neq i_{3} \neq i)$
     \For {$d=1 \to D$}
     \State $F=rand(0.1,1.5)$
     \State $\textbf{V}_{i,d}=\textbf{X}_{i_{1},d}+F(\textbf{X}_{i_{2},d}-\textbf{X}_{i_{3},d})$
     \EndFor
     
     \textit{//\textbf{End of Mutation}}
     
     \textit{//\textbf{Crossover}}
     
         \For {$d=1 \to D$}
            \If {$rand(0,1)<C_{r}$ or $d_{rand}=d$}
                   \State $U_{i,d}=V_{i,d}$
                   \Else
                   \State $U_{i,d}=x_{i,d}$      
            \EndIf
          \EndFor
           
\textit{//\textbf{End of Crossover}}  

\textit{//\textbf{Selection}}         
           
          \If {$f(\textbf{U}_{i}) \leq f(\textbf{X}_{i})$}
             \State $\textbf{X}^{\prime}_{i}=\textbf{U}_{i}$
          \Else
             \State  $\textbf{X}^{\prime}_{i}=\textbf{X}_{i}$
          \EndIf   
          
\textit{//\textbf{End of Selection}}          
          
  \EndFor
  \State $\textbf{X}_{i}={\textbf{X}}^{\prime}_{i}$, $\forall i\in\{1,...,N_{P}\}$
  \State $g=g+1$
\State $\textbf{P}^{g}=\{\textbf{X}_{1},...,\textbf{X}_{N_{P}}\}$
\EndWhile
\end{algorithmic}
\end{algorithm}
\alglanguage{pseudocode}

\begin{figure*}
        \begin{subfigure}[b]{0.5\textwidth}
        \centering
                \includegraphics[width=0.9\textwidth]{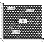}
                \caption{Mutation vector distribution of vector $R$}
                \label{fig:}
        \end{subfigure}%
         %add desired spacing between images, e. g. ~, \quad, \qquad etc.
        \begin{subfigure}[b]{0.5\textwidth}
        \centering
                \includegraphics[width=1.1\textwidth]{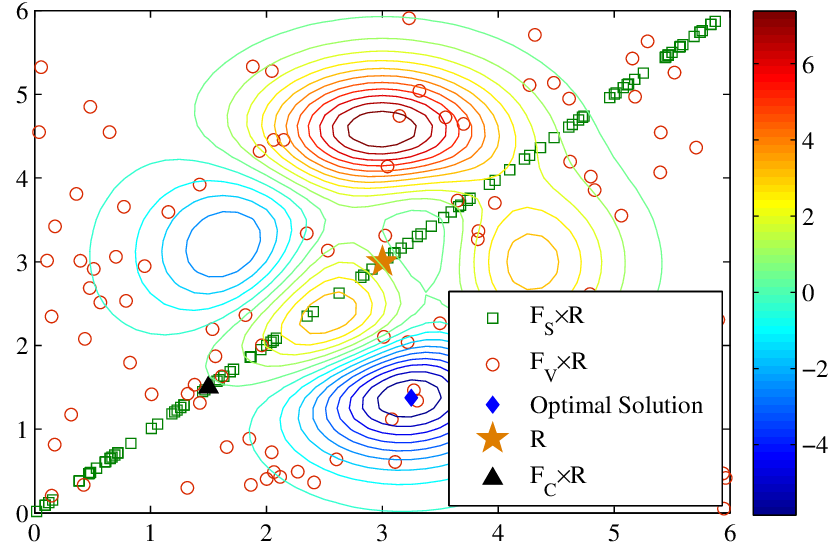}
                \caption{Monte-Carlo simulation}
                \label{fig:}
        \end{subfigure}
        \caption{Diversity of vector for a 2-D individual vector $R$ on a 2-D map for constant ($F_{C}$), scalar random ($F_{S}$), and vectorized random ($F_{V}$) mutation factors.}
        \label{fig:hex} 
\end{figure*}
\begin{figure*}
   \footnotesize
        \begin{subfigure}[b]{0.33\textwidth}
                \includegraphics[width=1\textwidth]{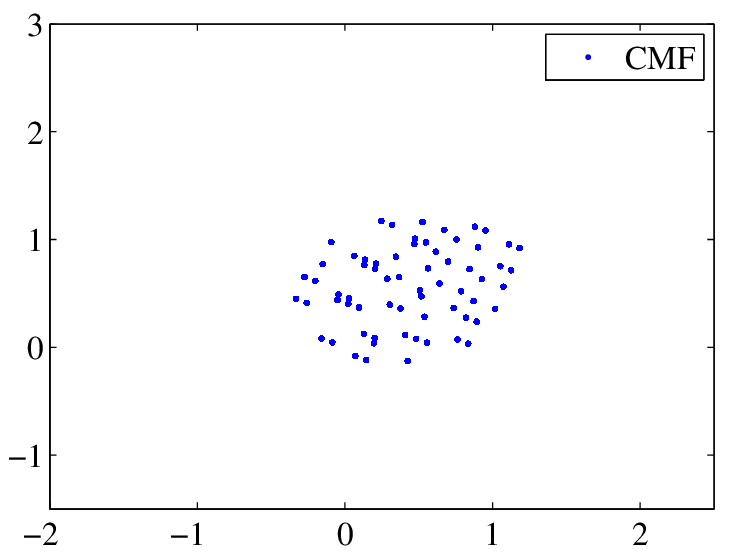}
                \caption{Static}
                \label{fig:}
        \end{subfigure}%
         %add desired spacing between images, e. g. ~, \quad, \qquad etc.
        \begin{subfigure}[b]{0.33\textwidth}
                \includegraphics[width=1\textwidth]{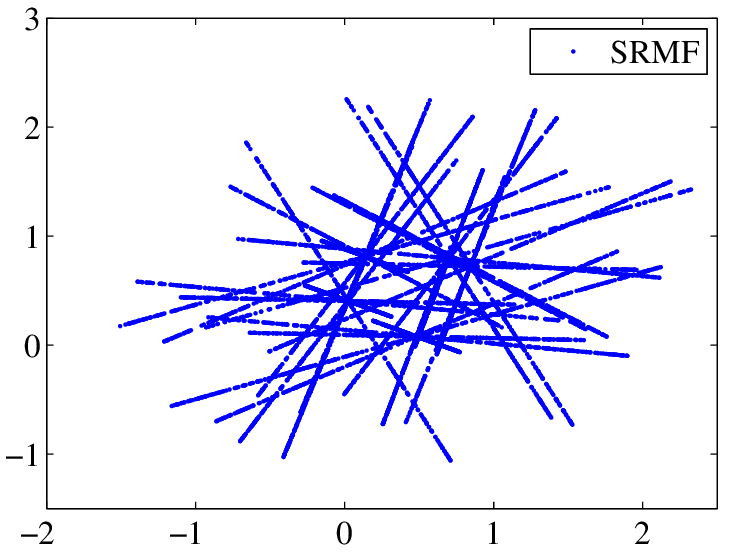}
                \caption{Scalar Random}
                \label{fig:}
        \end{subfigure}
  \begin{subfigure}[b]{0.33\textwidth}
                \includegraphics[width=1\textwidth]{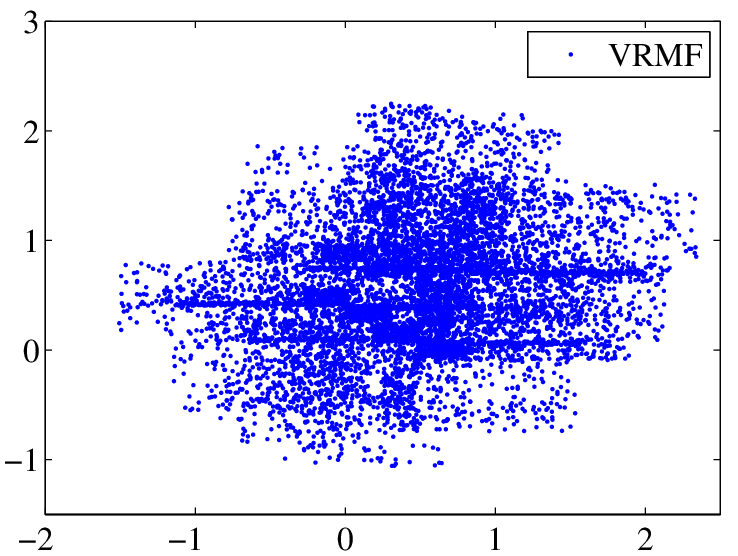}
                \caption{Vectorized Random} 
                \label{fig:}
        \end{subfigure}
\vspace{0.3cm}
        \caption{Monte-Carlo simulation of population diveristy for $D=2$ and $N_{P}=5$ after 10,000 random generation by considering the crossover operator.}\label{fig:monte3} 
\end{figure*}
\begin{figure*}
        \begin{subfigure}[b]{0.48\textwidth}
                \includegraphics[width=1\textwidth]{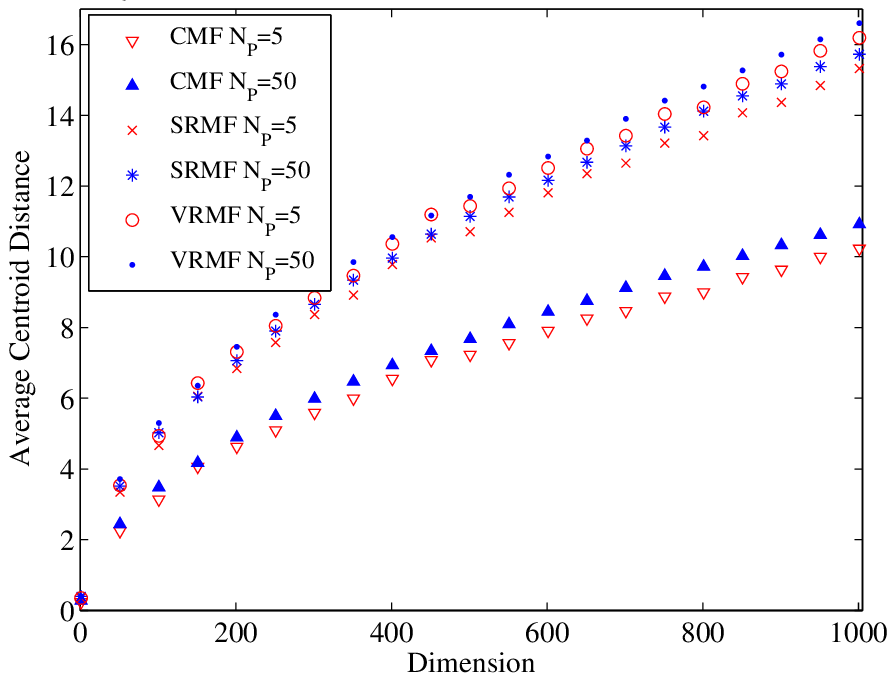}
                \caption{Centroid Distance}
                \label{fig:}
        \end{subfigure}%
         %add desired spacing between images, e. g. ~, \quad, \qquad etc.
        \begin{subfigure}[b]{0.48\textwidth}
                \includegraphics[width=1\textwidth]{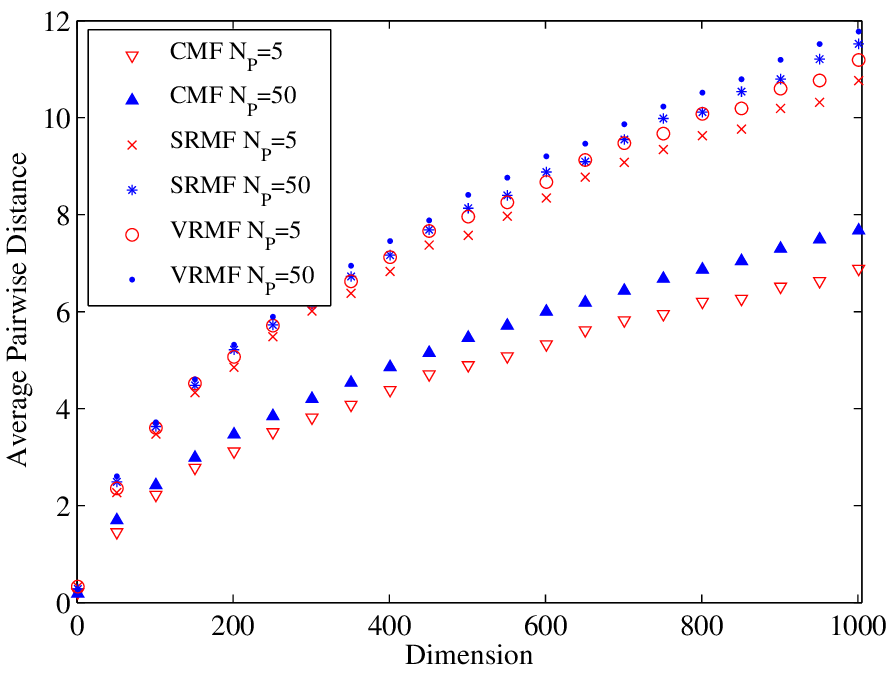}
                \caption{Pairwise Distance}
               
        \end{subfigure}
        \caption{Average centroid and pairwise distances for the Monte-Carlo simulation of population diversity for dimensions 1 to 1000 and $N_{P}\in\{5,50\}$ after 10000 random generations by considering the mutation and crossover operators.} \label{fig:distance}
\end{figure*}

The enhancement made in this paper, compared to the original idea proposed in \cite{39}, is parallel implementation of the MDEVM algorithm suitable for running on multicore central processing units (CPUs). In this implementation, the population is stored in the shared memory and a pool of workers (CPUs) is considered to conduct the processing. The procedure works in such a way that for each main step of the algorithm (Mutation, Crossover, Selection), the individuals are distributed over the CPU cores of the machines and reading/writing data is done in the  shared memory. In this way, we could enhance the running time of the algorithm dramatically, particularly suitable for application on smart devices. The pseudocode of the proposed MDEVM approach is in Algorithm 1.  After generation of initial population, the mutation vector is computed by using the proposed mutation factor, Eq. (\ref{Proposed_F}). Then, the \textit{crossover} and \textit{mutation} procedures are conducted similar to the DE algorithm to generate the next population. The termination criterion is met when the difference between best fitness value ($BFV$) and fitness value-to-reach ($VTR$) is less than fitness error-value-to-reach ($EVTR$), or the searching procedure exceeds the maximum number of function calls $NFC_{Max}$, i.e., $NFC\geq NFC_{Max}$. 

\subsection{Supporting Randomized Vectorized Mutation Factor By Monte-Carlo Simulations}
In order to visualize exploration abilities among CMF, scalar random mutation factor (SRMF), and VRMF, possible diversities of a 2-D individual sample vector $\textbf{R}$ is presented in Figure \ref{fig:hex}.a. In order to have a better sense of variable space, it is constructed with hexagons, where each hexagon represents a point on the variable space. The landscape for variables $x_{1}$ and $x_{2}$ is limited to boundaries $[a_{1},b_{1}]$ and $[a_{2},b_{2}]$. Therefore, by having the sample vector $\textbf{R}$, denoted with a dashed hexagon, effect of an arbitrary CMF on $\textbf{R}$ is denoted by $F_{C}\times\textbf{R}$, as a dotted dark hexagon. Therefore, diversity of the generated mutation vector $F_{C}\times\textbf{R}$ is limited to one hexagon (i.e. the dotted dark hexagon) on the direction of vector $\textbf{R}$. In the case of having an identical uniform random $F$ for all variables of an individual, i.e. the SRMF scheme, the diversity of mutation vector $F_{S}\times\textbf{R}$ is not just limited to one hexagon (i.e. the dotted dark hexagon), yet is along the vector $\textbf{R}$, denoted by grey hexagons. Conversely, by randomizing $F$ for each variable of each individual using a uniform random vector $\textbf{F}$, i.e. $F_{V}\times\textbf{R}$, the VRMF diversity covers the whole plane containing all the hexagons, which presents the highest exploration power.  

The diversities of CMF, SRMF, and VRMF are investigated by employing Monte-Carlo simulation on an arbitrary landscape in Figure \ref{fig:hex}.b. In this simulation for arbitrary vector $\textbf{R}=[1,1]$, 100 sample mutation vectors for each CMF, SRMF, and VRMF schemes with $F_{C}=0.5$ and $F_{S},F_{V}\in[0,2]$ are generated, where the variables are limited as $x_{1},x_{2}\in[0,3]$. The simulation illustrates that the VRMF scheme supports a higher diversity than the SRMF, where its diversity is limited to the points on a line. Strictly speaking, if all variables in the individual vector $\textbf{R}$ are multiplied by a random scalar number, other points are generated on the same direction of the line which is indicated by vector $F_{S}\times\textbf{R}$. In fact, the SRMF is generating points on the same direction as vector $\textbf{R}$. If the relationship among the variables (variables' interaction) are linear, the mutation vector is doing fine (which is a very exceptional case, especially during solving real-world problems). However, when the VRMF scheme is utilized, the mutation vector has no restriction to explore any point on the search space with no linearity restriction, which was the case for SRMF. This discussion is valid for higher dimensions, where the line needs to be replaced with a plane or hyperplane.

By taking into account the crossover component of MDE algorithm, another Monte-Carlo simulation is conducted for CMF, SRMF, and VRMF schemes as presented in Figure \ref{fig:monte3}. This simulations are conducted using the ``DE/Rand/1" mutation scheme for a population size of $N_{P}=5$, and $10000$ sample individuals are generated from an identical uniform random population, in a 2-D variables space, where each variable is uniform randomly selected as $x_{i}\in[0,1]$. The crossover plays a decisive role in taking diversity into the populations, as presented for the CMF scheme in Figure \ref{fig:monte3}.a. However as presented in Figure \ref{fig:monte3}.b and Figure \ref{fig:monte3}.c, the crossover also expands the diversity of SRMF and VRMF schemes dramatically such that almost the whole variable space is explored by the VRMF scheme. 

By keeping the stated Monte-Carlo simulation settings, the diversity analysis on CMF, SRMF, and VRMF schemes is extended for variable space dimensions $D\in\{1,...,1000\}$ and populations sizes $N_{P}\in\{5,50\}$ as shown in Figure \ref{fig:distance}. In these simulations, the average distance from the centroid point and pairwise distance measures are considered. The average distance from centroid demonstrated distance of each individual from the centroid of the population. This measure shows how diverse is the population. The average pairwise distance measure presents the average of distances between individual pairs in a population. This measure demonstrates the diversity of population as well as how far individuals are spreaded on the landscape from each other.

The average distance of individuals from the centroid distances is computed as:
\begin{equation}
C_{D}=\dfrac{1}{N_{P}}\sum\limits_{i=1}^{N_{P}} \sqrt{\sum\limits_{d=1}^{D}(x_{i,d}-x_{d}^{c})^2},
\end{equation}
where the centroid of the population is $\textbf{X}^{c}=(x_{1}^{c},...,x_{D}^{c})$, computed as
\begin{equation}
x_{d}^{c}=\dfrac{1}{N_{P}}\sum\limits_{i=1}^{N_{P}} x_{i,d},\:\forall d\in\{1,...,D\}.
\end{equation}
As Figure \ref{fig:distance}.a shows, the CMF has the least diversity for both $N_{P}=5$ and $N_{P}=50$ compared to SRMF and VRMF schemes. This is while the VRMF scheme has the highest diversity and as the dimensionality of problem increases, its diversity is improved more comparing to the CMF and SRMF schemes. It is obvious that the $N_{P}=50$ has a higher diversity than the $N_{P}=5$ in all schemes, but this diversity improvement is much less than the diversity that the VRMF scheme can deliver into the population with a much smaller population size, i.e. $N_{P}=5$. The comparison among CMF with $N_{P}=5$ and CMF with $N_{P}=50$ and VRMF with $N_{P}=5$ clearly indicates that the performance of VRMF scheme with small population size is higher in term of diversity enhancement. 

In order to study the diversity based on the average pairwise distance, it is computed as
\begin{equation}
P_{D}=\dfrac{1}{N_{P}(N_{P}-1)}\sum\limits_{i=1}^{N_{P}} \sum\limits_{\substack{j=1 \\ i \neq j}}^{N_{P}} \sqrt{\sum\limits_{d=1}^{D}(x_{i,d}-x_{j,d})^2}.
\end{equation}
The average pairwise distances for different dimensions and populations sizes $N_{P}=5$ and $N_{P}=50$ are illustrated in Figure \ref{fig:distance}.b. The simulation results for this diversity measurement criterion also clearly demonstrates strength of the VRMF with small populations size.

\begin{table}[t]
\caption{Parameter setting for all conducted experiments}
\begin{center}
\footnotesize
\begin{tabular}{|c|c|c|}
\hline
\multicolumn{1}{|c|}{Parameter}% time}
& \multicolumn{1}{|c|}{Description}% time}
& \multicolumn{1}{|c|}{Value} \\% cost}\\
 \hline
 \hline
$Cr$ & Crossover Probability Constant & 0.9\\ \hline
$NFC_{Max}$ & Maximum Number of Function Calls & $1000\times D$\\ \hline
$EVTR$ & Objective Function Error Value to Reach& 1e-8\\ \hline
$N_{Run}$ & Number of Runs & 30\\ \hline
$F$ & Mutation Factor & 0.9\\ \hline

\end{tabular}
\label{MDEVM_Parameter}
\end{center}
\end{table}

\section{Simulation Results}
In this section, performance of the proposed MDEVM algorithm is compared with the MDE, MDESM, and the $\mu$JADE \cite{Brown_2015} algorithms. The parameter setting and employed benchmark functions (i.e. CEC-2013 testbed \cite{19}) are described in the next subsection. Then, the comprehensive experimental series are presented in details. The algorithm is implemented in parallel using the multiprocessing library of Python programming language. The experiments are conducted on a cluster of 16 CPUs with 1TB of RAM. 

In the next subsection, the benchmark functions and parameters setting are provided. Afterwards, a set of experiments and analysis regarding different mutation schemes and population sizes, problem dimensionalities, mutation factor ranges, population diversity, and higher number of function calls is presented.

\begin{table}[!htp]
\caption{Mutant vector (MV) schemes for population sizes $N_{P}\in\{2,3,4\}$ and $N_{P}\geq5$.}
\begin{center}
\footnotesize
\begin{tabular}{|c|c|c|c|c|c|c|c|}

\hline
\multirow{1}{*}{$N_{P}$}&\multirow{1}{*}{MV}&{$\textbf{V}_{i}$}\\
%\hline
 \hline
\hline
\multirow{3}{*}{2}  & DE/Rand/1 & $\textbf{X}_{1}+F(\textbf{X}_{2})$ \\ \cline{2-3} 
                    & DE/Best/1 & $\textbf{X}_{best}+F(\textbf{X}_{1}-\textbf{X}_{2})$ \\ \cline{2-3} 
                    & DE/T2B/1 & $\textbf{X}_{i}+F(\textbf{X}_{best}-\textbf{X}_{i})+F(\textbf{X}_{1}-\textbf{X}_{2})$  \\ \hline

\multirow{3}{*}{3}  & DE/Rand/1 & $\textbf{X}_{1}+F(\textbf{X}_{2}-\textbf{X}_{3})$ \\ \cline{2-3} 
                    & DE/Best/1 & $\textbf{X}_{best}+F(\textbf{X}_{1}-\textbf{X}_{2})$ \\ \cline{2-3} 
                    & DE/T2B/1 & $\textbf{X}_{i}+F(\textbf{X}_{best}-\textbf{X}{i})+F(\textbf{X}_{1}-\textbf{X}_{2})$  \\ \hline

\multirow{3}{*}{4}  & DE/Rand/1 & $\textbf{X}_{1}+F(\textbf{X}_{2}-\textbf{X}_{3})$ \\ \cline{2-3} 
                    & DE/Best/1 & $\textbf{X}_{best}+F(\textbf{X}_{1}-\textbf{X}_{2})$ \\ \cline{2-3} 
                    & DE/T2B/1 & $\textbf{X}_{i}+F(\textbf{X}_{best}-\textbf{X}{i})+F(\textbf{X}_{1}-\textbf{X}_{2})$\\ \cline{2-3}  
                    & DE/Best/2 & $\textbf{X}_{best}+F(\textbf{X}_{1}-\textbf{X}_{2})+F(\textbf{X}_{3}-\textbf{X}_{4})$  \\ \hline
                    
\multirow{3}{*}{5$\leq$}& DE/Rand/1 & $\textbf{X}_{1}+F(\textbf{X}_{2}-\textbf{X}_{3})$ \\ \cline{2-3} 
                    & DE/Best/1 & $\textbf{X}_{best}+F(\textbf{X}_{1}-\textbf{X}_{2})$ \\ \cline{2-3} 
                    & DE/T2B/1 & $\textbf{X}_{i}+F(\textbf{X}_{best}-\textbf{X}{i})+F(\textbf{X}_{1}-\textbf{X}_{2})$\\ \cline{2-3}  
                    & DE/Best/2 & $\textbf{X}_{best}+F(\textbf{X}_{1}-\textbf{X}_{2})+F(\textbf{X}_{3}-\textbf{X}_{4})$  \\ \cline{2-3}  
                    & DE/Rand/2 & $\textbf{X}_{1}+F(\textbf{X}_{2}-\textbf{X}_{3})+F(\textbf{X}_{4}-\textbf{X}_{5})$  \\ \hline                  
\end{tabular}
\label{MDEVM_Mutation Vectors}
\end{center}
\end{table}

\begin{table}[!htp]
\caption{Number of Wilcoxon rank-sum test comparisons for MDEVM against MDE, MDESM, and $\mu$JADE schemes on CEC 2013 benchmark functions and population sizes $N_{P}\in\{2,3,4,5,6,50\}$ for dimension $D=50$ and mutation vector (MV) schemes ``DE/Rand/1", ``DE/Best/1", ``DE/T2B/1", ``DE/Best/2", and ``DE/Rand/2". If the bolded value is under ``+'' column, the MDEVM method has the highest overall performance, otherwise, the corresponding method under the column header has the best overall performance.}
\begin{center}
\footnotesize
\begin{tabular}{|c|c|c|c|c|c|c|c|c|c|c|}

\hline
\multirow{2}{*}{$N_{P}$}&\multirow{2}{*}{MV}&\multicolumn{3}{c|}{MDE}&\multicolumn{3}{c|}{MDESM}&\multicolumn{3}{c|}{$\mu$JADE}\\
\hhline{~~---------}
& &+&=&-&+&=&-&+&=&-\\
%\hline
\hhline{-----------}
\hhline{-----------}

\multirow{3}{*}{2}  & DE/Rand/1 & 0 & 23 & \textbf{5} & 2 & 19 & \textbf{7} &\textbf{2} &25 &1 \\ \cline{2-11} 
                    & DE/Best/1 & \textbf{14} & 11 & 3 & \textbf{15} & 9 & 4&\textbf{15} &4 &9 \\ \cline{2-11} 
                    & DE/T2B/1 & \textbf{25} & 3 & 0 & \textbf{17} & 9 & 2 & \textbf{13}&10 &5 \\ \hhline{-----------}
\multirow{3}{*}{3}  & DE/Rand/1 & \textbf{11} & 10 & 7 & 7 & 11 & \textbf{10}& \textbf{12}& 13& 3\\ \cline{2-11} 
                    & DE/Best/1 & \textbf{24} & 4 & 0 & \textbf{20} & 5 & 3& \textbf{8}&15 &5 \\ \cline{2-11} 
                    & DE/T2B/1 & \textbf{17} & 9 & 2 & \textbf{12} & 10 & 6 &6 & 15&\textbf{7} \\ \hhline{-----------}
                    
\multirow{4}{*}{4}  & DE/Rand/1 & \textbf{20} & 4 & 4 & \textbf{12} & 5 & 11& \textbf{14}& 1& 13\\ \cline{2-11} 
                    & DE/Best/1 & \textbf{21} & 7 & 0 & \textbf{19} & 5 & 4&\textbf{12} &5 &11 \\ \cline{2-11} 
                    & DE/T2B/1 & \textbf{20} & 4 & 4 & \textbf{17} & 1 & 10& \textbf{15}&1 &12 \\ \cline{2-11} 
                    & DE/Best/2 & 2 & 3 & \textbf{23} & 0 & 4 & \textbf{24} & 0&3 &\textbf{25} \\ \hhline{-----------}
                    
\multirow{5}{*}{5}  & DE/Rand/1 & \textbf{19} & 2 & 7 & \textbf{12} & 8 & 8& \textbf{14}& 2& 12\\ \cline{2-11} 
                    & DE/Best/1 & \textbf{24} & 2 & 2 & \textbf{16} & 7 & 5& \textbf{15}&1 &12 \\ \cline{2-11} 
                    & DE/T2B/1 & \textbf{19} & 2 & 7 & \textbf{15} & 4 & 9&\textbf{18} &3 &7 \\ \cline{2-11} 
                    & DE/Best/2 & \textbf{12} & 6 & 10 & 6 & 7 & \textbf{15}&10 &2 &\textbf{16} \\ \cline{2-11} 
                    & DE/Rand/2 & 0 & 1 & \textbf{27} & 1 & 1 & \textbf{26}&4 &4 &\textbf{20}  \\ \hhline{-----------}
                    
\multirow{5}{*}{6}  & DE/Rand/1 & \textbf{13} & 5 & 10& 13 & 7 & \textbf{8}&10 & 4&\textbf{14}  \\ \cline{2-11} 
                    & DE/Best/1 & \textbf{21} & 3 & 4 & \textbf{18 }& 6 & 4& \textbf{14}& 3&11\\ \cline{2-11} 
                    & DE/T2B/1 & \textbf{19} & 5 & 4 & \textbf{15 }& 2 & 11& \textbf{14}&2 & 12\\ \cline{2-11} 
                    & DE/Best/2 & 11 & 5 & \textbf{12} & 7 & 4 & \textbf{17}& 8& 9& \textbf{11}\\ \cline{2-11} 
                    & DE/Rand/2 & 1 & 1 & \textbf{26} & 1 & 2 & \textbf{25} &1 &8 &\textbf{19} \\ \hhline{-----------}

\multirow{5}{*}{50} & DE/Rand/1 & 1 & 2 & \textbf{25} & 1 & 3 &  \textbf{24}& 2& 3& \textbf{23} \\ \cline{2-11} 
                    & DE/Best/1 & 10 &  5 &  \textbf{13} & 2 & 6 &  \textbf{20}&2 &4 &\textbf{22} \\ \cline{2-11} 
                    & DE/T2B/1 & 0 & 3 & \textbf{25} & 0 & 4 &  \textbf{24}&0 &4 & \textbf{24} \\ \cline{2-11} 
                    & DE/Best/2 & 0 & 4 & \textbf{ 24} & 0 & 3 &  \textbf{25}& 0&2 & \textbf{26}\\ \cline{2-11} 
                    & DE/Rand/2 & 0 & 2 & \textbf{26} & 1 & 2 &\textbf{25} &0 &1 &\textbf{27} \\ \hhline{-----------}
                    \hhline{-----------}
 %\multicolumn{2}{|c|}{Number of Successes}   & \textbf{14} & - & 11& 12 & - & \textbf{13}&\textbf{11}& -&14 \\ \cline{2-11} 
 %\hhline{-----------}
 %\hhline{-----------}                                  
%\multirow{2}{*}{Summary} &  MDEVM &  \multicolumn{3}{c|}{MDE}  &  \multicolumn{3}{c|}{MDESM} &  \multicolumn{3}{c|}{$\mu$JADE}  \\ \cline{2-11} 
% \hhline{~----------}
%  \multirow{1}{*}{} &  \%52 & \multicolumn{3}{c|}{\%16} &  \multicolumn{3}{c|}{\%08} &  \multicolumn{3}{c|}{\%24}  \\ \cline{2-11} 
% \hhline{-----------}
                            
\end{tabular}
\label{MDEVM_AllRes}
\end{center}
\end{table}

\subsection{Benchmark Functions and Parameters Setting}
All the experiments are conducted on the CEC-2013 testbed \cite{19}. It is comprised of $28$ benchmark functions and an improved version of CEC-2005 \cite{20} counterpart with additional test functions and modified formula in order to create the composite functions, oscillations, and symmetric-breaking transforms. This testbed is divided into three categories which are uni-modal functions ($f_{1}-f_{5}$), multi-modal functions ($f_{6}-f_{20}$), and composite functions ($f_{21}-f_{28}$) \cite{19}.
Parameters setting for all the experiments are presented in Table ~\ref{MDEVM_Parameter} adapted from the literature \cite{3}, \cite{14}, \cite{19}, unless a change is mentioned. The reported values are averaged for $N_{Run}=30$ independent runs per function per algorithm to minimize the effect of the stochastic nature of the algorithms on the reported results. 

The mutation schemes presented by Eq.(1) to Eq.(5) are the five main schemes, which are used for $N_{P}\geq5$ in experiments \cite{61}, \cite{62}. For the small size and very small size populations, we are using the mutation schemes based on their structure for different sizes of population as demonstrated in Table~\ref{MDEVM_Mutation Vectors}. For the $N_{P}=2$, we have proposed a ``DE/Rand/1'' mutation vector scheme as
\begin{equation}
\textbf{X}_{1}+F(\textbf{X}_{2})
\end{equation}
where the only two available individuals in the population are used in it. We have used the ``DE/Rand/1", ``DE/Best/1", ``DE/T2B/1", and ``DE/Best/2" schemes for $N_{P}<4$.

\subsection{Experimental Series 1: Mutation Schemes and Population Size Analysis}
Performance of the MDE, MDESM, MDEVM, and $\mu$JADE schemes are evaluated for mutation schemes in Table \ref{MDEVM_Mutation Vectors}, population sizes $N_{P}\in\{2,3,4,5,6,50\}$, and dimension $D=50$. The Wilcoxon test results are reported in terms of pair-wise comparison in Table \ref{MDEVM_AllRes}. The symbols ``+", ``=", and ``-" indicate a statistically better, equivalent, and worse performance, respectively, compared with the MDEVM algorithm \cite{63}. 

\begin{figure*}[!htp]
   \centering    
   \footnotesize
        \includegraphics[scale=0.65]{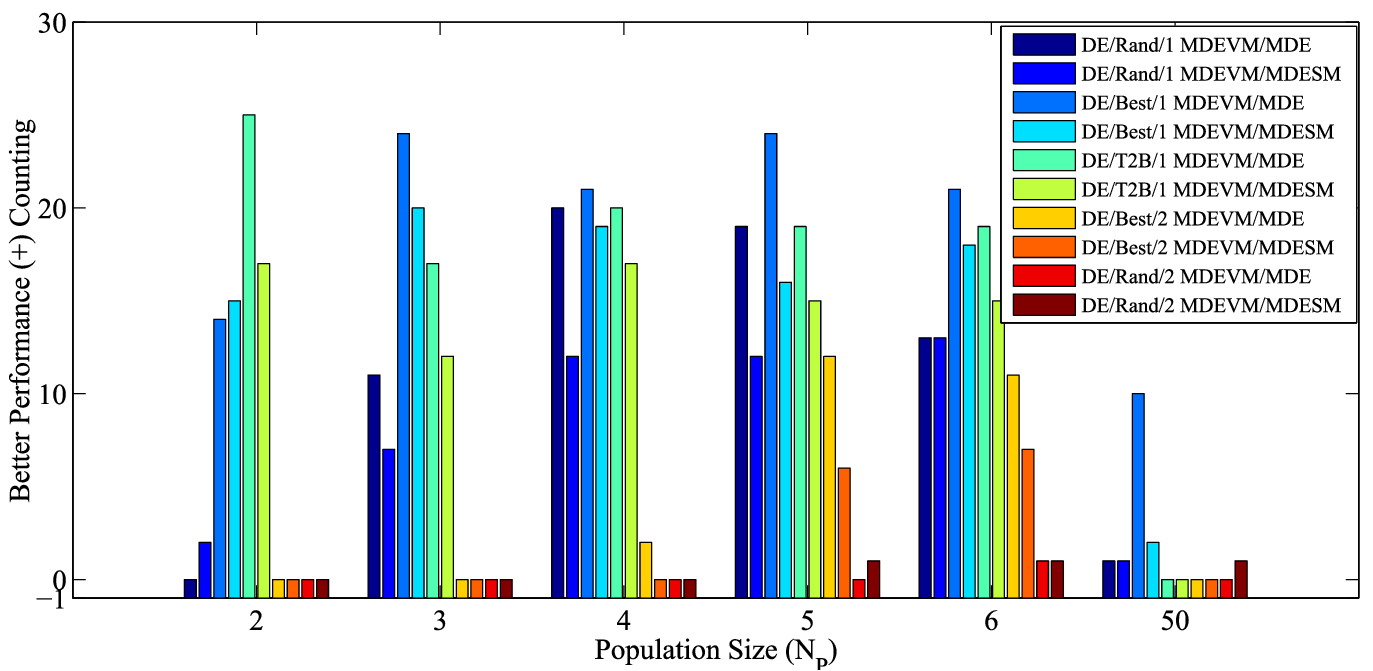}
        \caption{The better (+) performance counting for the MDEVM vs. MDE and MDEVM vs. MDESM comparison for different mutation schemes and populations sizes.}
        \label{fig:allbars}     
\end{figure*}

\begin{figure*}
\footnotesize
        \begin{subfigure}[b]{0.31\textwidth}
           \centering 
                 \includegraphics[width=1\linewidth]{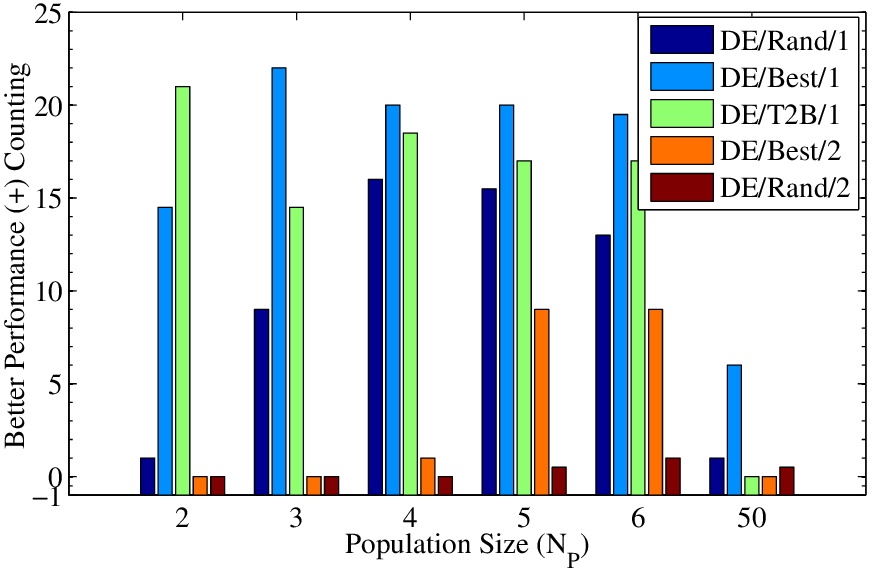}
                \caption{Better (+) performance.}
                \label{fig:}
        \end{subfigure}%       
\vspace{0.2cm}
 \begin{subfigure}[b]{0.31\textwidth}
    \centering 
                 \includegraphics[width=1\linewidth]{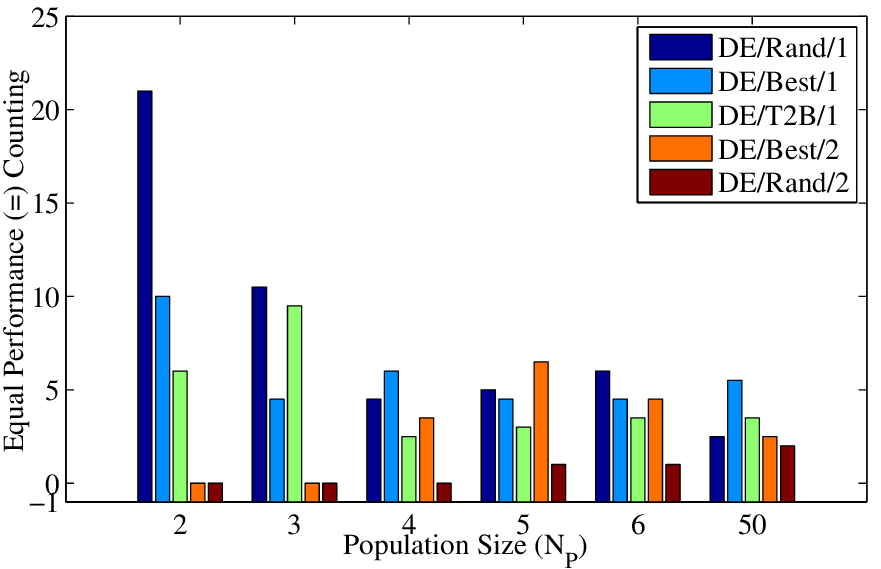}
                \caption{Equal (=) performance.}
                \label{fig:}
        \end{subfigure}%
\vspace{0.2cm}
 \begin{subfigure}[b]{0.31\textwidth}
    \centering 
                 \includegraphics[width=1\linewidth]{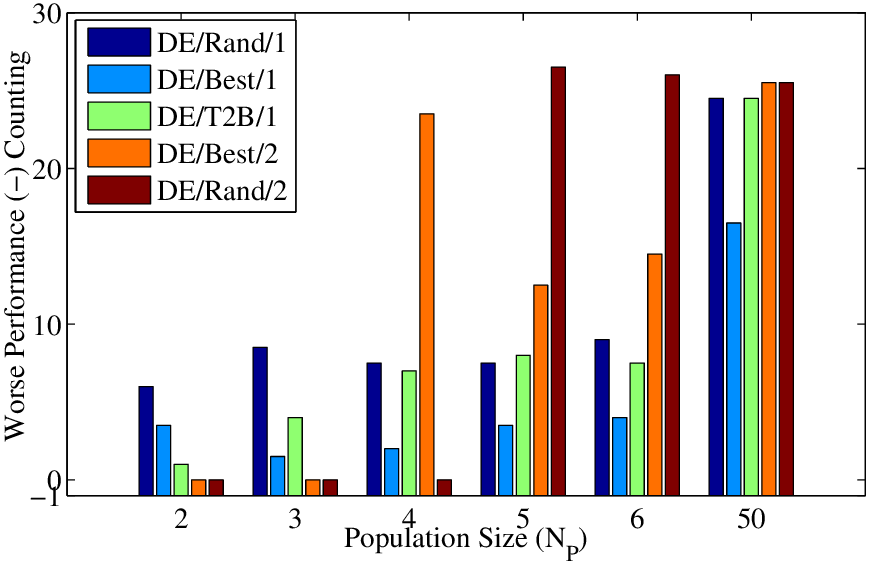}
                \caption{Worse (-) performance.}
                \label{fig:}
        \end{subfigure}%
\vspace{0.2cm}
        \caption{Average performance of MDEVM vs. MDE and MDEVM vs. MDESM methods for different mutation schemes and populations sizes.}\label{fig:} 
                \label{fig:threebars}     
\end{figure*}
\begin{figure*}
        \includegraphics[scale=0.55]{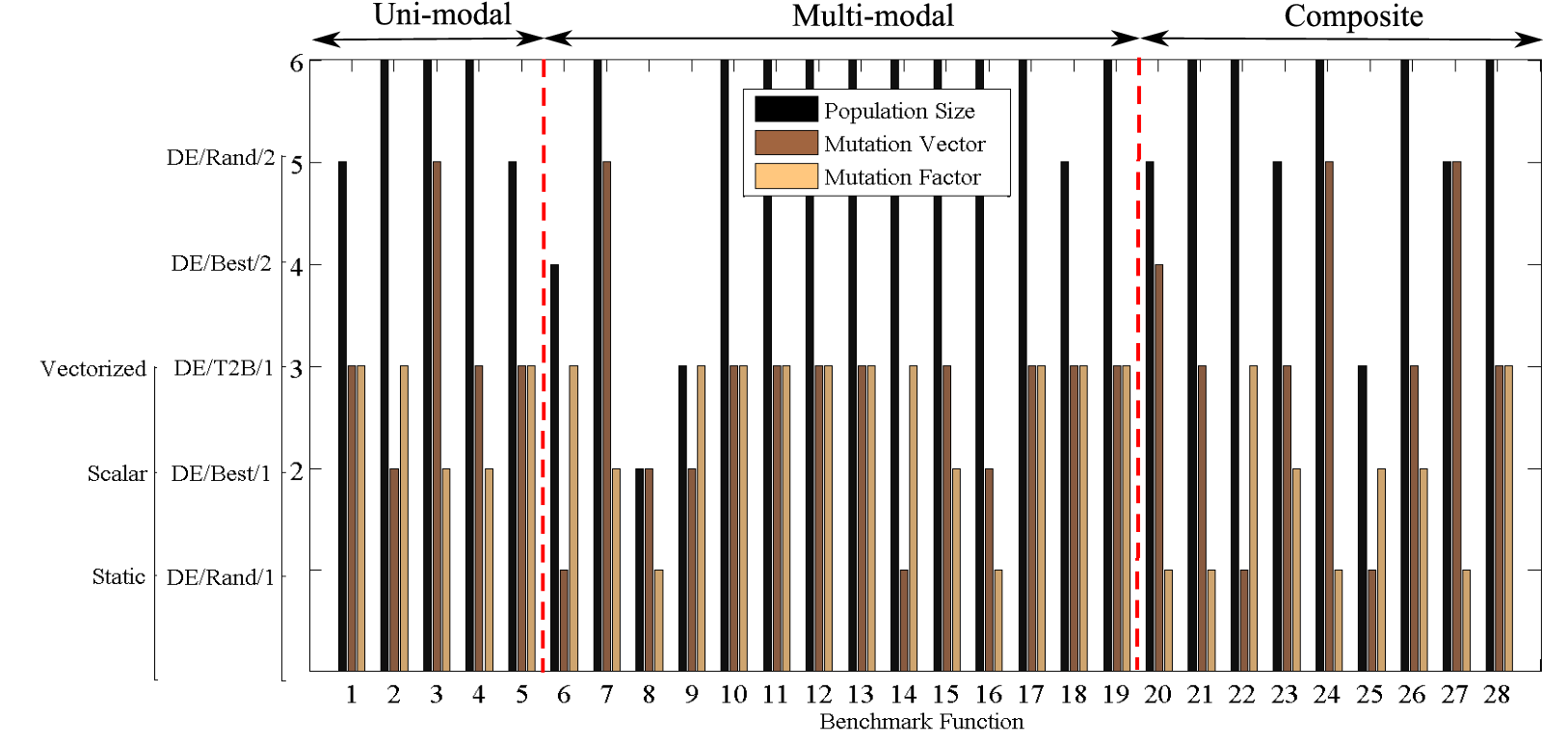}
        \caption{Components of highest performance algorithm with respect to the best error for each benchmark function, and function families uni-modal ($f_{1}-f_{5}$), multi-modal ($f_{6}-f_{19}$), and composite ($f_{20}-f_{28}$).}
        \label{fig:functionbar}     
\end{figure*}

The results in Table~\ref{MDEVM_AllRes} demonstrate that generally the MDEVM method the $\mu$JADE method have competitive results. 
It is interesting that for very small population size (i.e. $N_{P}\leq5$), the MDEVM method has competitive or better performance than other methods. This is particularly obvious for the ``DE/Best/1'' and ``DE/T2B/1'' schemes with $N_{P}=2$ and ``DE/Best/1'' scheme for $N_{P}=3$. Regarding $N_{P}\in{4,5}$, we see that the MDEVM method has more successful results for the ``DE/Rand/1'', ``DE/Best/1'', and ``DE/T2B/1'' schemes. However, as the diversity of population is increasing by adding more number of individuals to the population, the MDEVM method achieves less successful results. This situation is obvious for $N_{P}=50$ where a standard size of population is used and we see the $\mu$JADE, MDE, and MDESM methods have much better performance. The results clearly show that for the $N_{P}\geq5$, the VRMF technique in MDEVM can add a good diversity to the population which results in better performance that other methods. However, this diversity enhancement method has extra affect on large population sizes, in a way that the population has more than enough diversity and cannot converge to an optima, which is the stagnation situation. Since the ``DE/Best/2" and ``DE/Rand/2" schemes have more exploration capability due to incorporating more population individuals, using the VRMF technique results in extra diversity in the population. This is another additive diversity which stops the MDEVM method to converge to optimal solution(s). The difference between DE and MDE algorithms is in population size which delivers diversity into the population. Combining the VRMF technique with the DE algorithm consequences in extra diversity which results a poor performance of the algorithm. Using the standalone DE-algorithm may result a better performance, but by the cost of more number of function calls. Therefore, utilizing the MDE algorithm with small population sizes can deliver both higher diversity and performance into the algorithm. In overall, the ``DE/Best/1", ``DE/Rand/1", and ``DE/T2B/1" schemes have the best performance among the various mutation schemes for MDEVM.

In Figure~\ref{fig:allbars}, a summary of better performance counting of all schemes is presented, where $N_{P}=5$ has the highest number of success for all mutation schemes on average. In order to have a closer look, average of better, equal, and worse performance counting for the MDEVM vs. MDE and MDEVM vs. MDESM comparisons are presented in Figure~\ref{fig:threebars}. 

Regarding the average of better and equivalent performances results as shown in Figure~\ref{fig:threebars}.a and Figure~\ref{fig:threebars}.b, it is clear that the ``DE/Best/1" scheme has the most number of successes. In terms of worse performance comparison, it is interesting that as the population size increases, the number of worse performance counts, particularly for the ``DE/T2B/1", ``DE/Best/2", and ``DE/Rand/2" mutation schemes, increase dramatically. 

\begin{figure*}[!htp]
        \begin{subfigure}[b]{0.33\textwidth}
                 \includegraphics[width=1\linewidth]{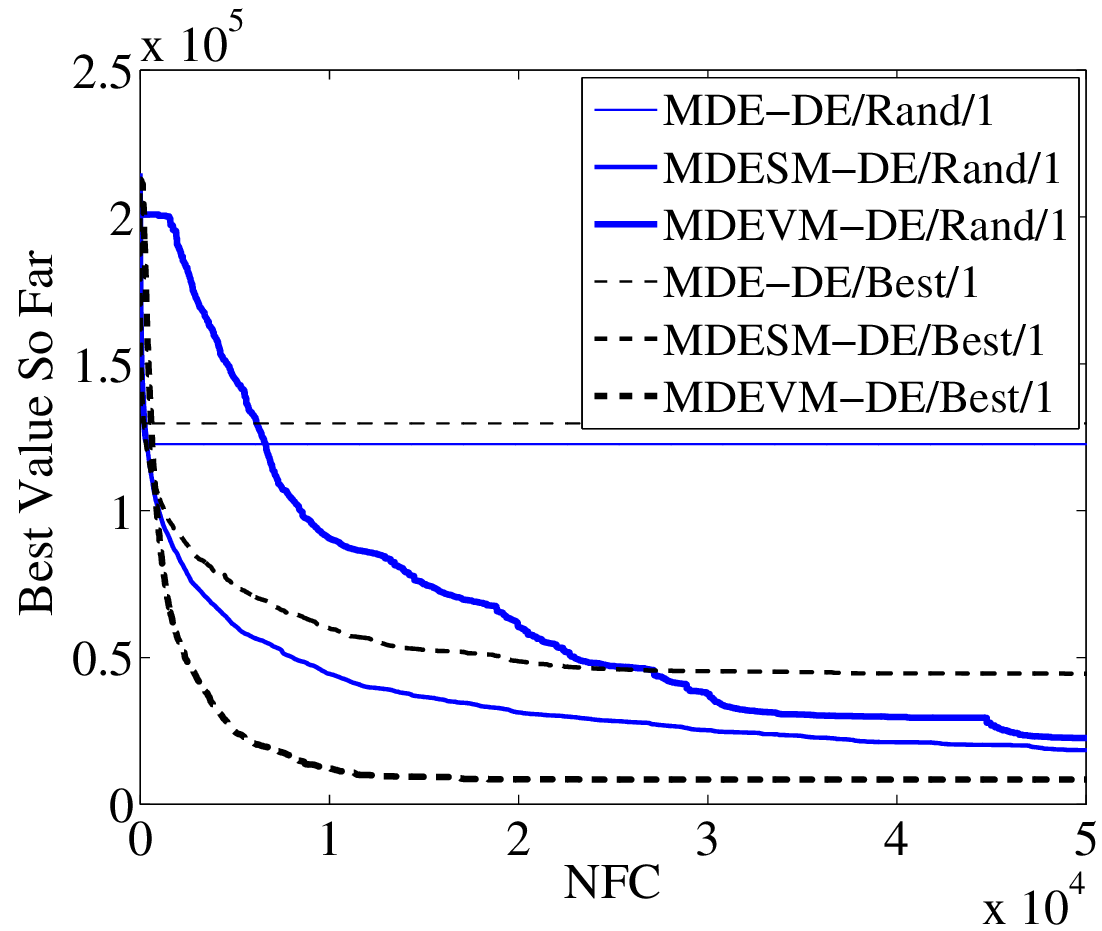}
                \caption{$f_{1}$}
                \label{fig:1}
        \end{subfigure}%       
% \begin{subfigure}[b]{0.33\textwidth}
%                 \includegraphics[width=1\linewidth]{sF2.eps}
%                \caption{$f_{2}$}
%                \label{fig:}
%        \end{subfigure}%
% \begin{subfigure}[b]{0.33\textwidth}
%                 \includegraphics[width=1\linewidth]{sF3.eps}
%                \caption{$f_{3}$}
%                \label{fig:}
%        \end{subfigure}%
%        
%        \vspace{0.2cm}
% \begin{subfigure}[b]{0.33\textwidth}
%                 \includegraphics[width=1\linewidth]{sF4.eps}
%                \caption{$f_{4}$}
%                \label{fig:}
%        \end{subfigure}%        
%        \begin{subfigure}[b]{0.33\textwidth}
%                 \includegraphics[width=1\linewidth]{sF5.eps}
%                \caption{$f_{5}$}
%                \label{fig:}
%        \end{subfigure}%       
 \begin{subfigure}[b]{0.33\textwidth}
                 \includegraphics[width=1\linewidth]{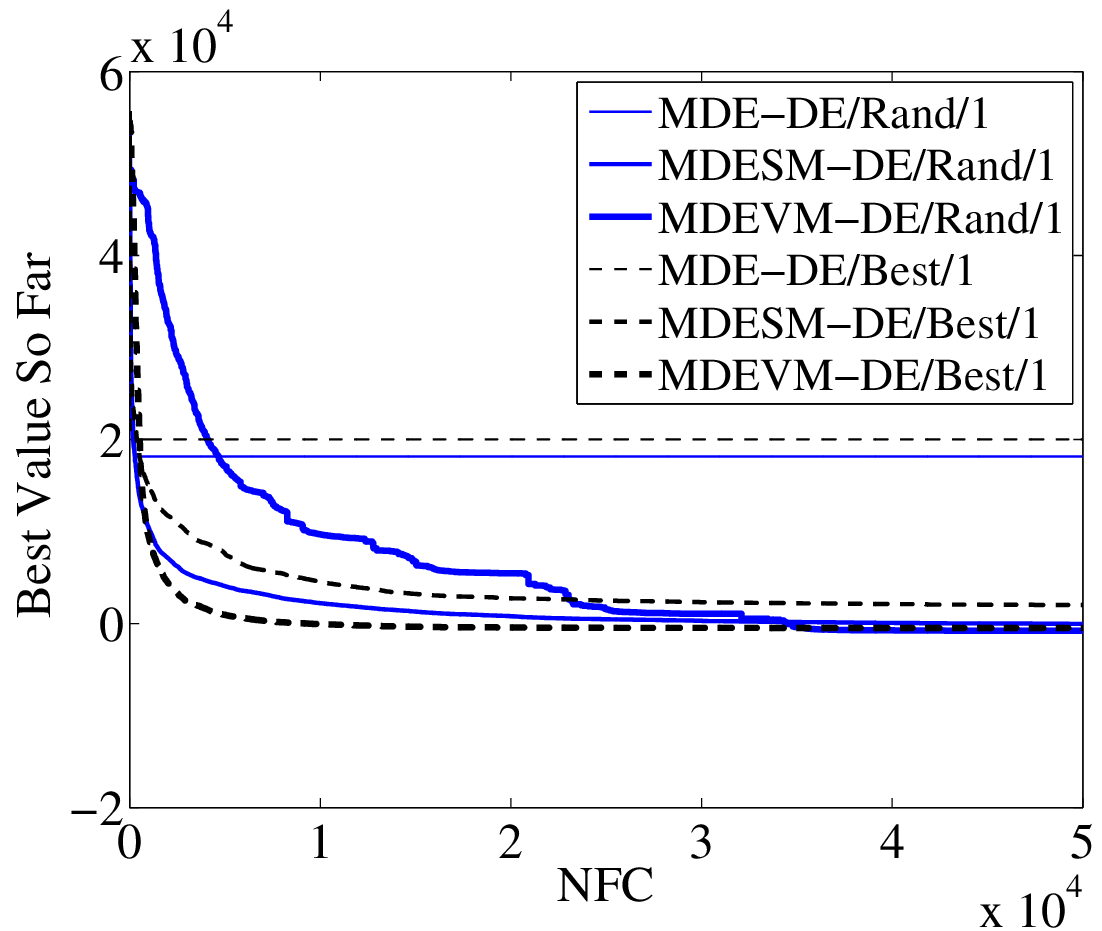}
                \caption{$f_{6}$}
                \label{fig:}
        \end{subfigure}%       
 \begin{subfigure}[b]{0.33\textwidth}
                 \includegraphics[width=1\linewidth]{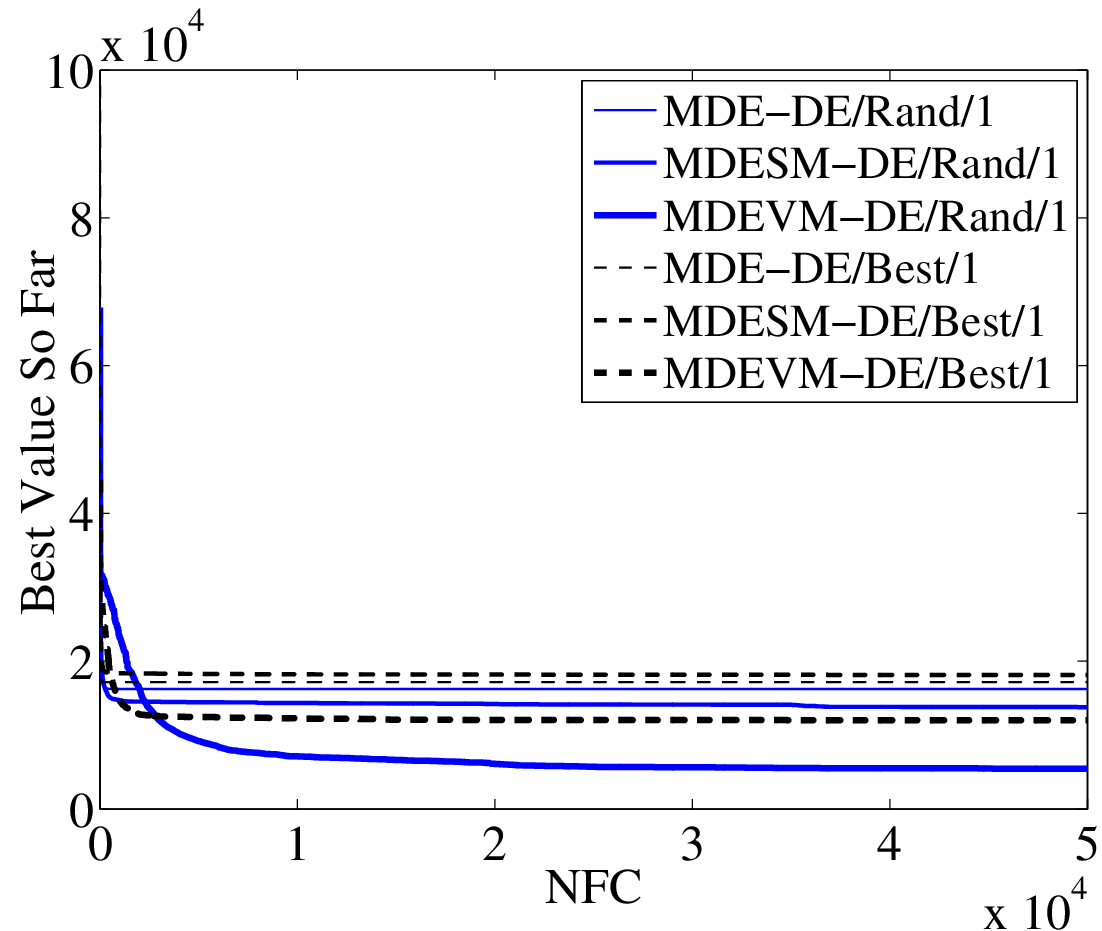}
                \caption{$f_{28}$}
                \label{fig:}
        \end{subfigure}%               
        \caption{Best value so far of the MDE, MDESM, and MDEVM schemes for the DE/Rand/1 and DE/Best/1 and $N_{P}=5$. For bravity, some functions are selected.}\label{fig:np5} 
\end{figure*}

% \caption{Average of MDEVM/MDE and MDEVM/MDESM preformances for differenet mutation schemes and populations sizes.}\label{fig:} 
%\end{figure*}     
%   \begin{figure*} [!htp]       

The best error value for each benchmark function family is illustrated in Figure~\ref{fig:functionbar}. The dash line separates the uni-modal, multi-modal, and composite, benchmark functions types. For the uni-modal and multi-modal functions, the VRMF method with the ``DE/T2B/1" mutation scheme and $N_{P}=6$ has the best performance. For the composite functions, the SRMF method with the ``DE/T2B/1" mutation scheme and $N_{P}=6$ has the best performance. In overall, the ``DE/Best/1" mutation scheme with population size of $N_{P}=5$ is recommended as the well-performance scheme among the all. Further analysis are conducted on ``DE/Best/1" scheme in deep, including the popular scheme ``DE/Rand/1".
  
In Figure~\ref{fig:np5}, performance of the MDE, MDESM, and MDEVM methods for the ``DE/Rand/1" and ``DE/Best/1" mutation schemes and different number of function calls are presented. As an example for the $f_{1}$ in Figure~\ref{fig:1}, the MDEVM method with the ``DE/Best/1" has converged faster and the MDEVM method with the ``DE/Rand/1" is going to converge with a sharp slope. By assigning a higher number of possible function calls, this method can outperform the MDEVM method with the ``DE/Best/1" scheme. The algorithms for different number of function calls are discussed further in the current section. Similar behaviour as above is obvious for $f_{14}$, $f_{20}$ and $f_{22}$. Such behaviour is due to the natural diversity in the ``DE/Rand/1" scheme. 

\begin{table}[t]
\caption{Number of Wilcoxon rank-sum test comparisons for MDEVM vs. MDE, MDESM, and $\mu$JADE methods on CEC 2013 benchmark functions and population size $N_{P}=5$ for dimension $D\in\{10,30,50,100\}$ and mutation vector (MV) schemes ``DE/Rand/1" and ``DE/Best/1". If the bolded value is under ``+'' column, the MDEVM method has the highest overall performance, otherwise, the corresponding method under the column header has the best overall performance.}
\begin{center}
\footnotesize
\begin{tabular}{|c|c|c|c|c|c|c|c|c|c|c|}

\hline
\multirow{2}{*}{$D$}&\multirow{2}{*}{MV}&\multicolumn{3}{c|}{MDE}&\multicolumn{3}{c|}{MDESM}&\multicolumn{3}{c|}{$\mu$JADE}\\
\hhline{~~---------}
& &+&=&-&+&=&-&+&=&-\\
%\hline
 \hline
\hline

\multirow{2}{*}{10}  & DE/Rand/1 & \textbf{23}&	3&	2	&\textbf{12}&	13	&3& \textbf{13}& 3&12 \\ \cline{2-11} 
                    & DE/Best/1 &\textbf{26}	&0	&2	&\textbf{24}&	3&	1& \textbf{14}&4 &10\\ \hline 

\multirow{2}{*}{30}  & DE/Rand/1 & \textbf{22}&	4	&2&	\textbf{17}&	5&	6& \textbf{14}& 2&12\\ \cline{2-11} 
                    & DE/Best/1 & \textbf{21}&	5	&2&	\textbf{20}	&6&	2& \textbf{15}& 1&12\\ \hline
                    
\multirow{2}{*}{50}  & DE/Rand/1 & \textbf{19} & 2 & 7 & \textbf{12} & 8 & 8&11 &4 &\textbf{13}\\ \cline{2-11} 
                    & DE/Best/1 & \textbf{24} & 2 & 2 & \textbf{16} & 7 & 5&10 &5 &\textbf{13} \\ \hline
                    
\multirow{2}{*}{100}  & DE/Rand/1& \textbf{18} & 3 & 7 & \textbf{16} & 5 & 7 &\textbf{11} &7 &10\\ \cline{2-11} 
                    & DE/Best/1 & \textbf{20} & 5 & 3 & \textbf{15 }& 9 & 4& 10&6 &\textbf{12}\\  \hline 

 %\multicolumn{2}{|c|}{Number of Successes}   & 11 & 0 & 4 & 0 & 0 & 2& 2& 0&6 \\ \cline{2-11} 
 %\hhline{-----------}
 %\hhline{-----------}                                  
%\multirow{2}{*}{Summary} &  MDEVM &  \multicolumn{3}{c|}{MDE}  &  \multicolumn{3}{c|}{MDESM} &  \multicolumn{3}{c|}{$\mu$JADE}  \\ \cline{2-11} 
% \hhline{~----------}
%  \multirow{1}{*}{} &  \%52 & \multicolumn{3}{c|}{\%16} &  \multicolumn{3}{c|}{\%08} &  \multicolumn{3}{c|}{\%24}  \\ \cline{2-11} 
% \hhline{-----------}
                                      
\end{tabular}
\label{D_analysis}
\end{center}
\end{table}

\subsection{Experimental Series 2: Dimensionality Effects}
In this subsection, performance of the proposed MDEVM method is compared with the MDE, MDESM, and $\mu$JADE methods for dimension $D\in\{10,30,50,100\}$ and population size $N_{P}=5$ with ``DE/Rand/1" and ``DE/Best/1" mutation vector schemes regarding the best value so far value. By considering the MDEVM method as the reference algorithm, summary of the Wilcoxon test results are reported in terms of pair-wise comparisons in Table~\ref{D_analysis}. The results clearly demonstrate that the proposed MDEVM method has outperformed the MDE and MDESM methods for different dimensions. The MDESM method shows a better performance than the MDE method, which is due to the SRMF diversity enhancement technique used in this scheme. Both ``DE/Rand/1" and ``DE/Best/1" mutation schemes have competitive performances over all dimensions and MDE schemes. As the dimensionality of problems increases, the $\mu$JADE method provides competitive performance versus the MDEVM method. This shows that for high-dimensional problems, the adaptive method along with a small size of population can provide a good diversity. We can see the same situation with the MDEVM method where the diversification of the mutation factor for decision variables can help the small size population to increase its diversity, suitable for high-dimensional problems.  

\begin{table}[t]
\caption{Summary of performance results of the MDESM and MDEVM approaches with $F\in[0,2]$ and $\textbf{F}\in[0,2]$, respectively, versus the MDESM and MDEVM with $F\in[0.1,1.5]$ and $\textbf{F}\in[0.1,1.5]$, respectively. The second set of methods are denoted with index $F$. $N_{P}=5$ and $D=30$.}
\begin{center}
\footnotesize
\begin{tabular}{|c|c|c|c|c|c|c|c|}

\hline
\multirow{2}{*}{Method}&\multirow{2}{*}{MV}&\multicolumn{3}{c|}{MDESM$_{F}$}&\multicolumn{3}{c|}{MDEVM$_{F}$}\\
\hhline{~~------}
&&+&=&-&+&=&-\\
%\hline
 \hline
\hline
\multirow{2}{*}{MDESM} &DE/Rand/1&14&12&2&6&7&15\\ \cline{2-8} 
 &DE/Best/1&14&14&0&7&9&12\\ \hline
\multirow{2}{*}{MDEVM} & DE/Rand/1&19&5&4&3&25&0\\ \cline{2-8} 
& DE/Best/1&19&6&3&19&3&6\\ \hline                                
\end{tabular}
\label{F_analysis}
\end{center}
\end{table}

\subsection{Experimental Series 3: Mutation Factor's Range Analysis}
The most common mutation factor in the literature is $F=0.5$, selected from the recommended range $F\in[0,2]$, \cite{62}. Recently, different values for $F$ and its range has been proposed, such as $F=0.7$ in \cite{1} and $F\in[0.1,1.5]$ in \cite{3}. Therefore, some experiments are conducted in this subsection to analyse affect of mutation factor range, on the performance of the MDESM and MDEVM approaches. By considering $N_{P}=5$ for dimension $D=30$ and mutation vector schemes ``DE/Rand/1" and ``DE/Best/1", the best error, standard deviation, and Wilcoxon rank-sum test results by considering the MDESM and MDEVM algorithms as references are presented in Table~\ref{F_analysis}. The mutation factor ranges are considered as $F\in[0,2]$ and $\textbf{F}\in[0,2]$ for MDESM and MDEVM approaches. The approaches with the range $F\in[0.1,1.5]$ and $\textbf{F}\in[0.1,1.5]$ are denoted by index $F$, which are MDESM$_{F}$ and MDEVM$_{F}$. The $W_{S}$ and $W_{V}$ demonstrate performance of the MDESM and MDEVM methods versus the MDESM$_{F}$ and MDEVM$_{F}$ methods, respectively. 

As demonstrated in Table \ref{F_analysis}, the MDESM method has almost better performance than the MDESM$_{F}$ method. However, the MDEVM$_{F}$ method has outperformed the MDESM method due to the delivered diversity by the VRMF approach into the MDEVM$_{F}$ method. The results of comparing MDEVM with the MDESM$_{F}$ and MDEVM$_{F}$ methods demonstrate that selecting $\textbf{F}$ in the interval $[0,2]$ has a better performance than the limited interval $[0.1,1.5]$. The comparison between the MDEVM and MDEVM$_{F}$ also shows almost equal performance. Overall, better performance of the MDEVM method is obvious, since the MDEVM method has diversity served from both VRMF and wider mutation factor range $[0,2]$. 

\subsection{Experimental Series 4: Population's Diversity Analysis}

\begin{figure*} [!htp]  
\small
     \begin{subfigure}[b]{0.32\textwidth}
                 \includegraphics[width=1\linewidth]{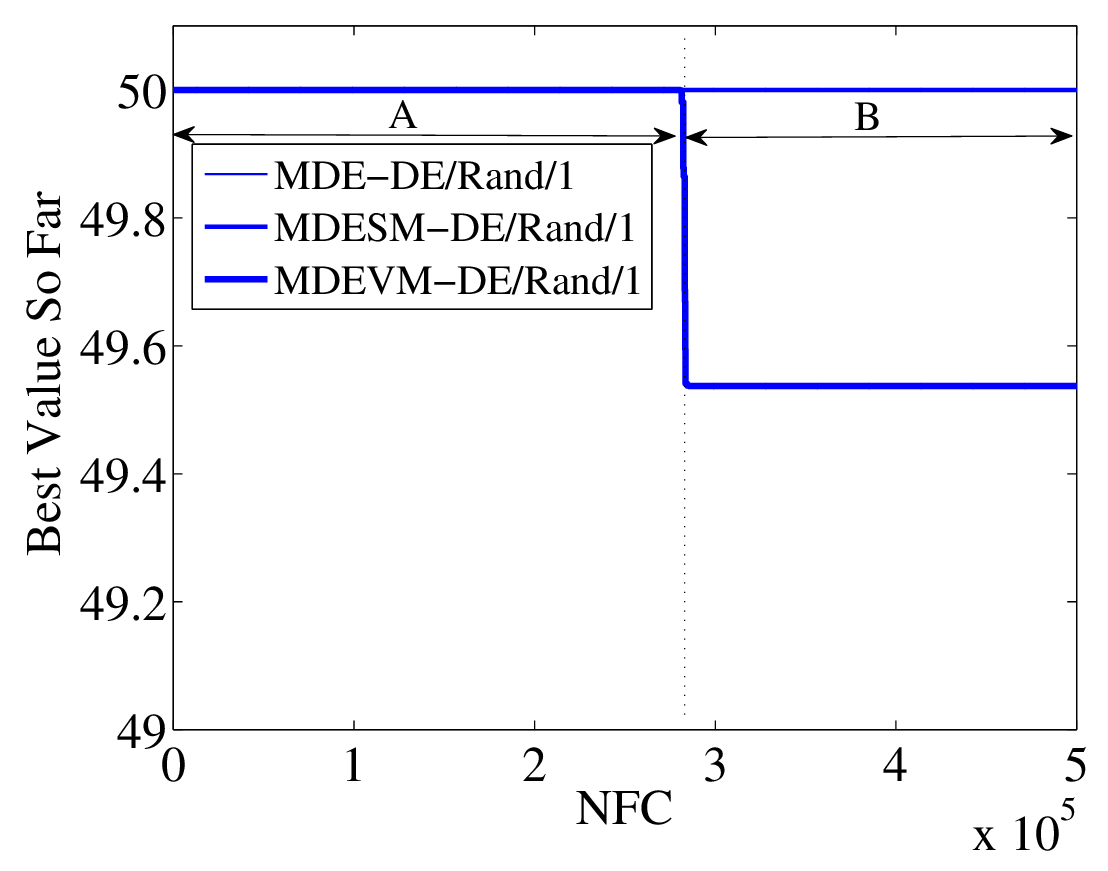}
                \caption{$f_{20}$: DE/Rand/1 Scheme}
                \label{HNF20}
        \end{subfigure}%    
         \begin{subfigure}[b]{0.32\textwidth}
                 \includegraphics[width=1\linewidth]{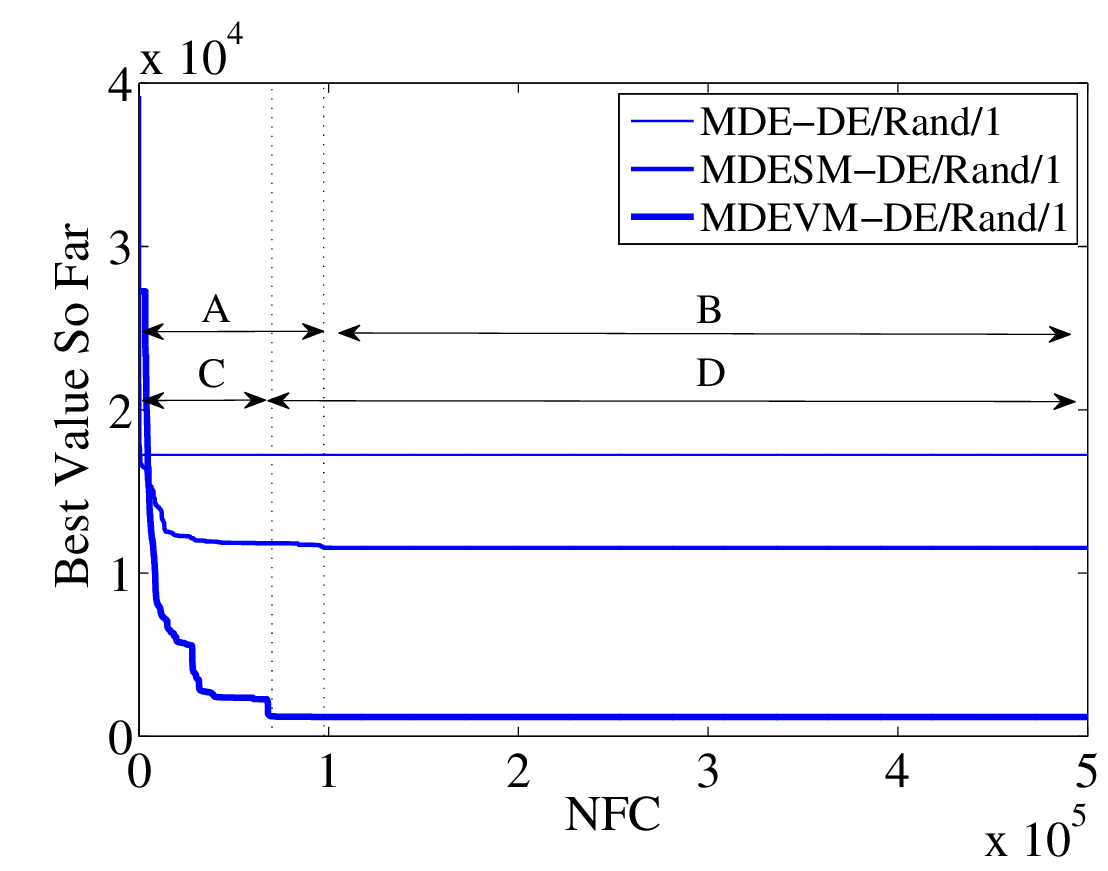}
                \caption{$f_{21}$: DE/Rand/1 Scheme}
                \label{fig:}
        \end{subfigure}%
        \begin{subfigure}[b]{0.32\textwidth}
                 \includegraphics[width=1\linewidth]{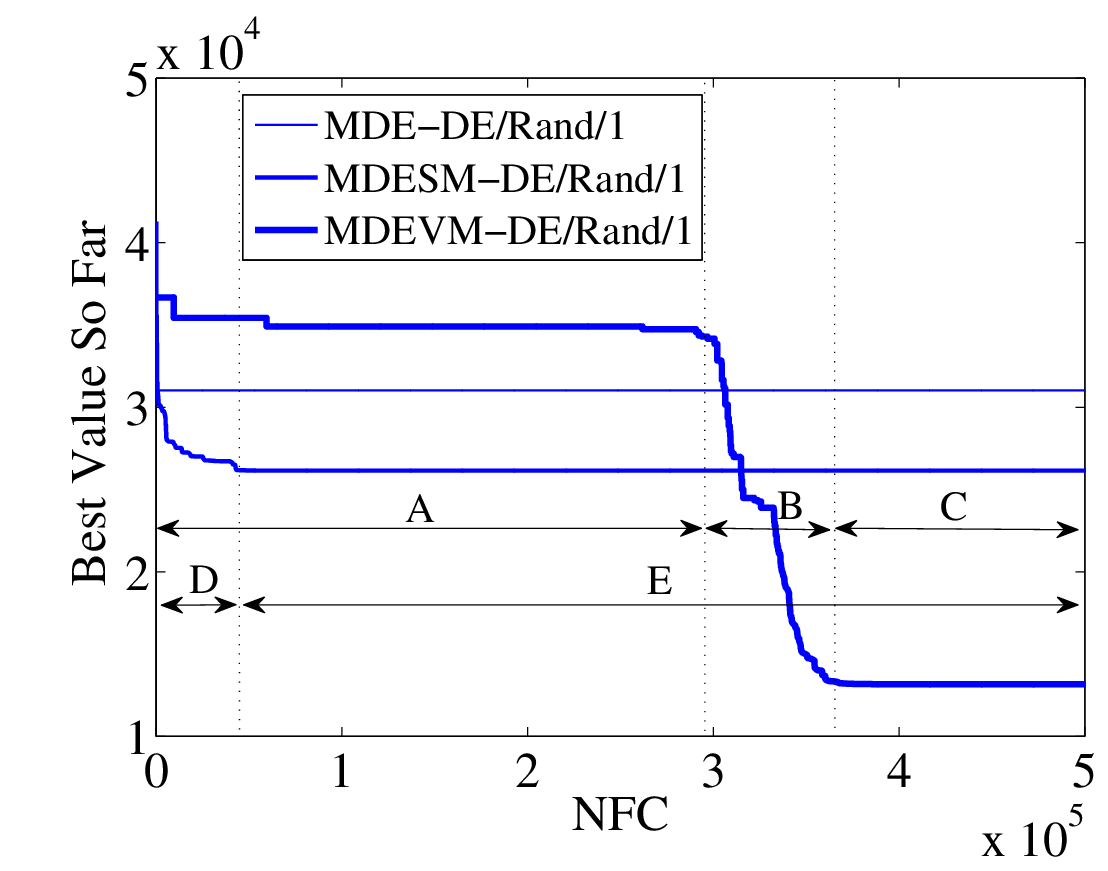}
                \caption{$f_{22}$: DE/Rand/1 Scheme}
                \label{HNF20}
        \end{subfigure}%    
    
         \begin{subfigure}[b]{0.32\textwidth}
                 \includegraphics[width=1\linewidth]{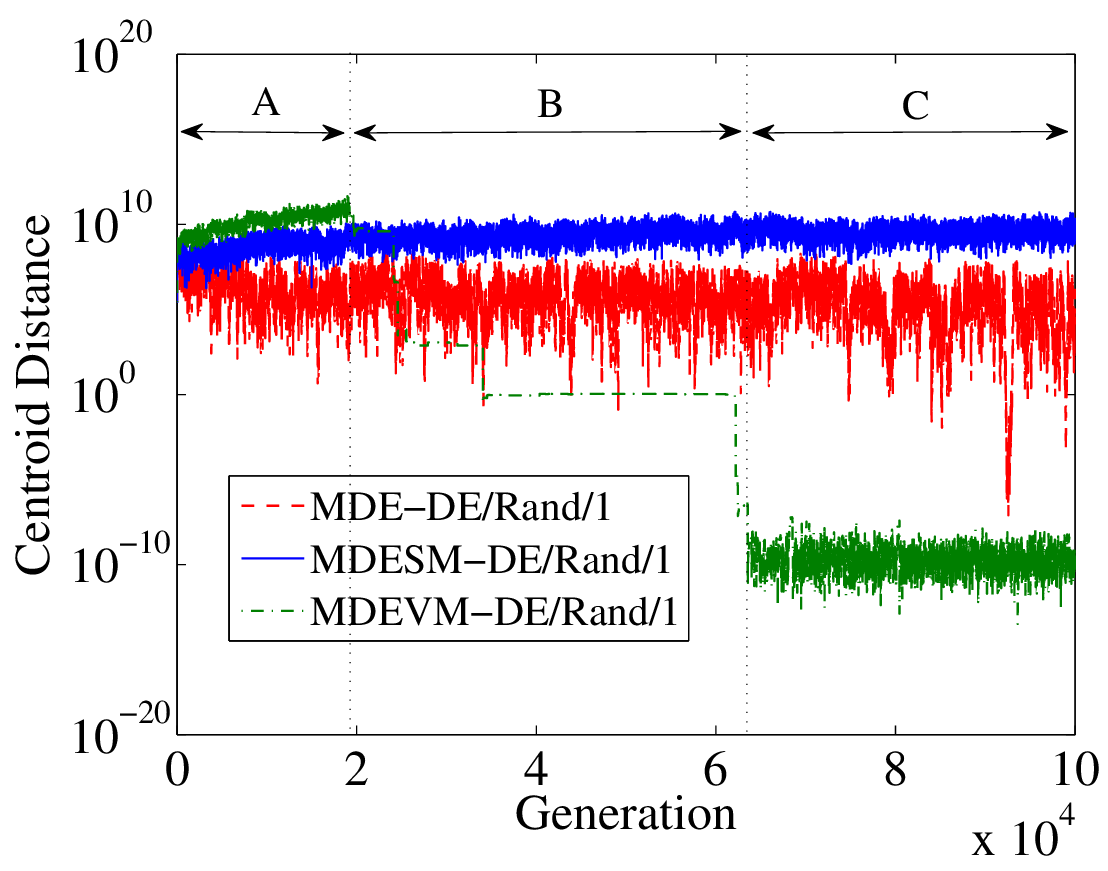}
                \caption{$f_{20}$: DE/Rand/1 Scheme}
                \label{HNF20}
        \end{subfigure}%    
         \begin{subfigure}[b]{0.32\textwidth}
                 \includegraphics[width=1\linewidth]{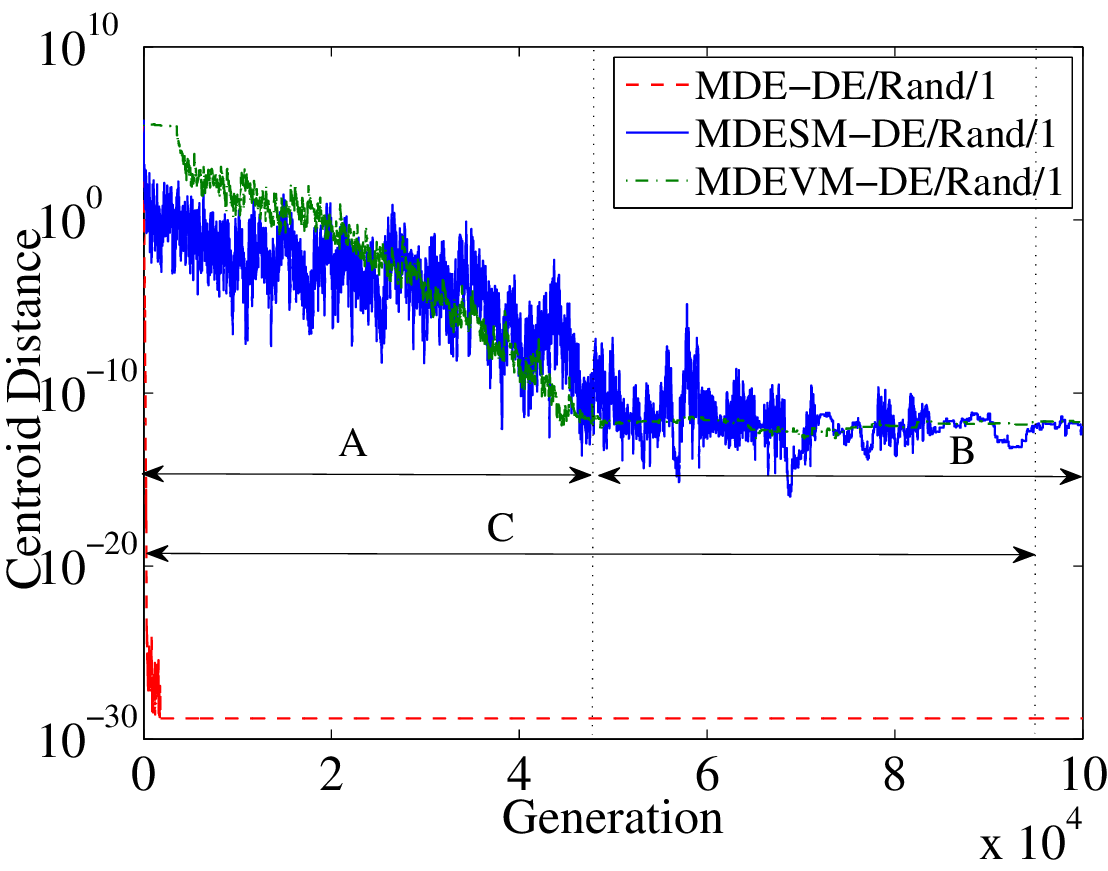}
                \caption{$f_{21}$: DE/Rand/1 Scheme}
                \label{fig:}
        \end{subfigure}%
        \begin{subfigure}[b]{0.32\textwidth}
                 \includegraphics[width=1\linewidth]{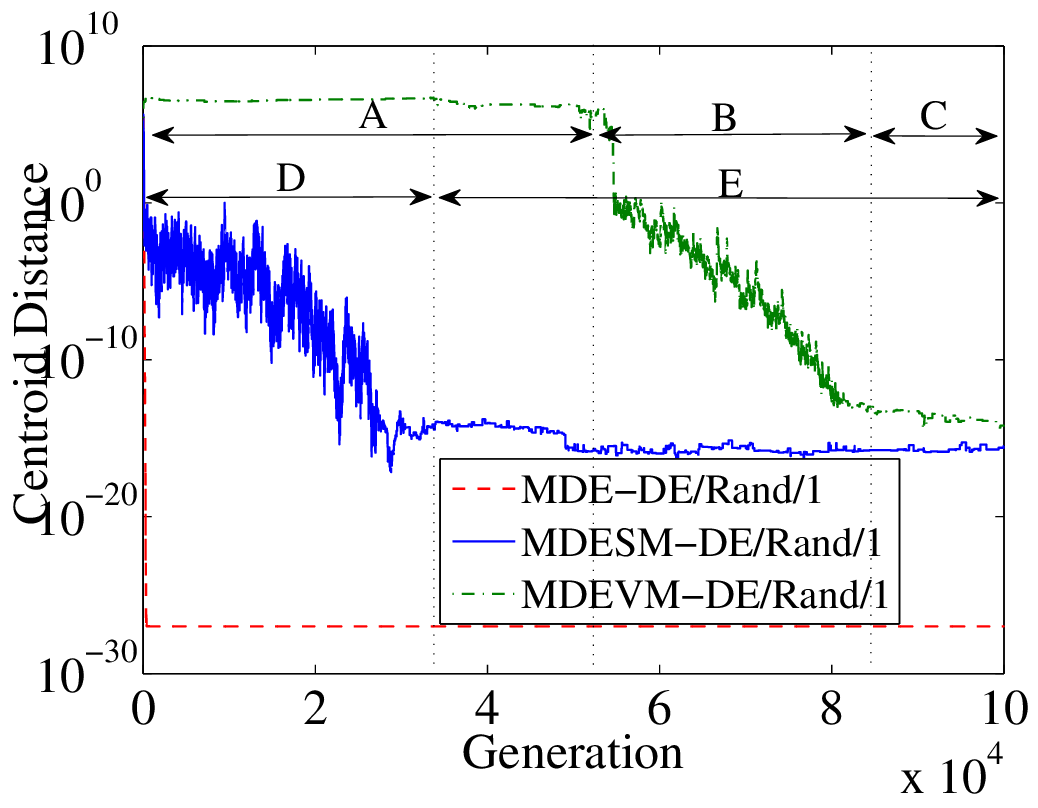}
                \caption{$f_{22}$: DE/Rand/1 Scheme}
                \label{HNF20}
        \end{subfigure}%
          
           \begin{subfigure}[b]{0.32\textwidth}
                 \includegraphics[width=1\linewidth]{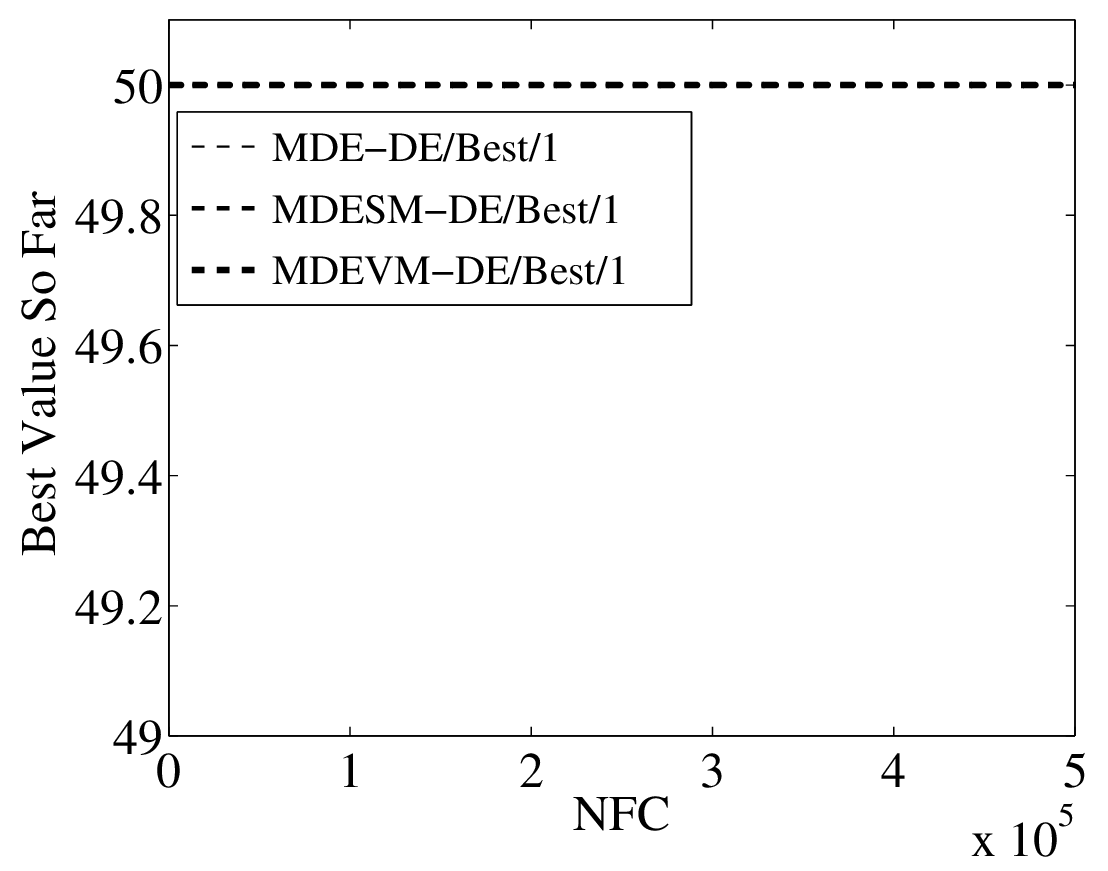}
                \caption{$f_{20}$: DE/Best/1 Scheme}
                \label{HNF20}
        \end{subfigure}%    
         \begin{subfigure}[b]{0.32\textwidth}
                 \includegraphics[width=1\linewidth]{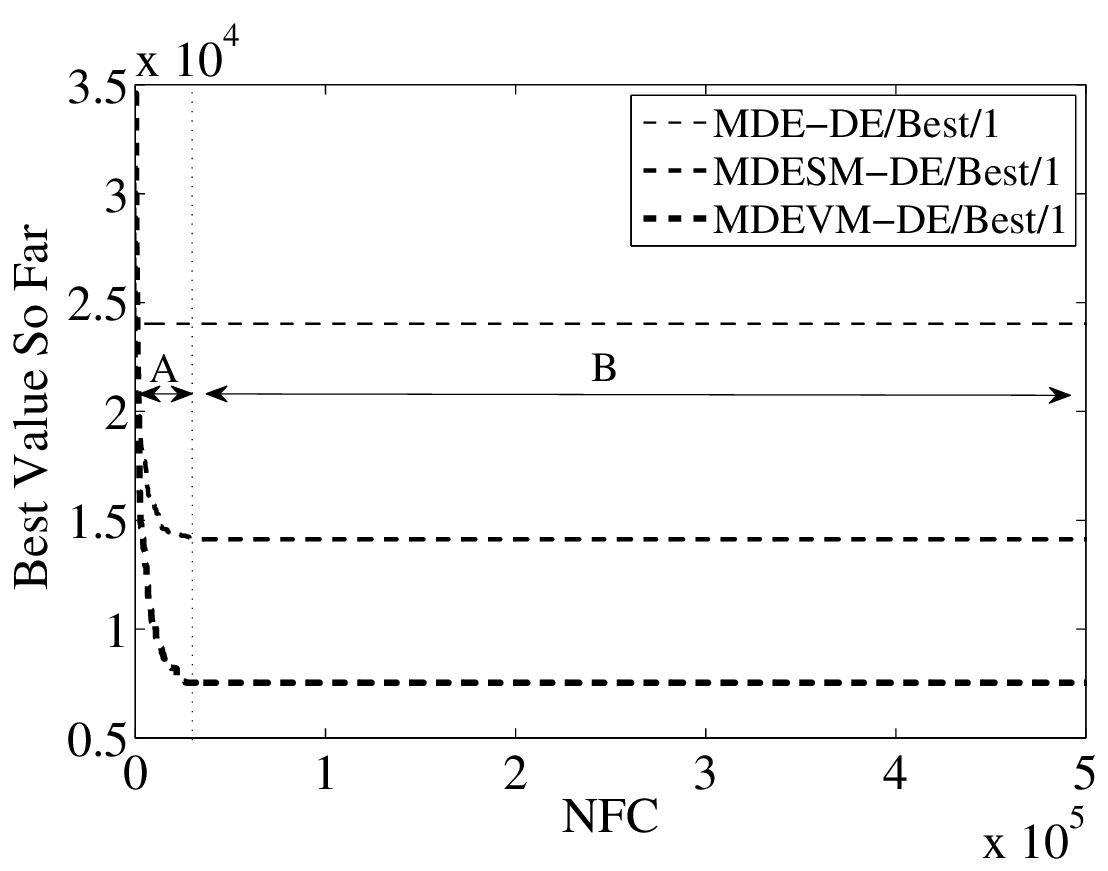}
                \caption{$f_{21}$: DE/Best/1 Scheme}
                \label{fig:}
        \end{subfigure}%
        \begin{subfigure}[b]{0.32\textwidth}
                 \includegraphics[width=1\linewidth]{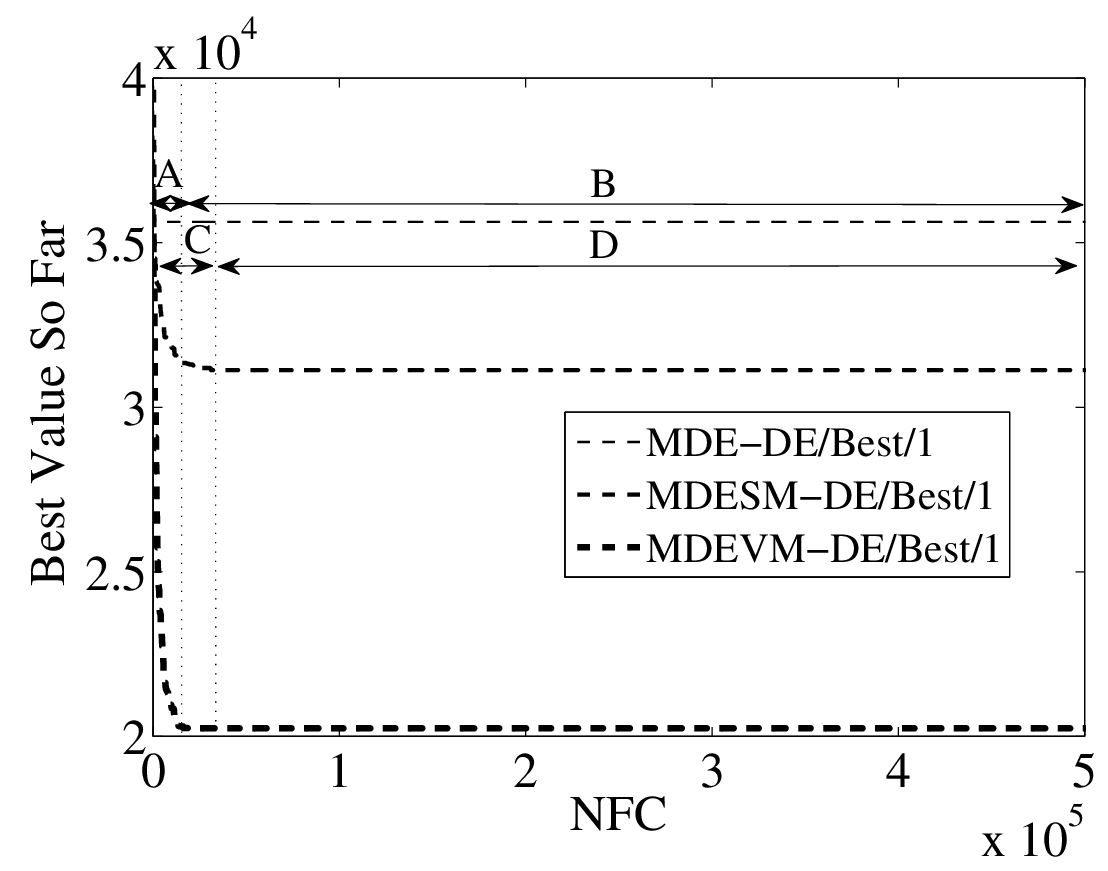}
                \caption{$f_{22}$: DE/Best/1 Scheme}
                \label{HNF20}
        \end{subfigure}%
        
  \begin{subfigure}[b]{0.32\textwidth}
                 \includegraphics[width=1\linewidth]{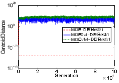}
                \caption{$f_{20}$: DE/Best/1 Scheme}
                \label{HNF20}
        \end{subfigure}%    
         \begin{subfigure}[b]{0.32\textwidth}
                 \includegraphics[width=1\linewidth]{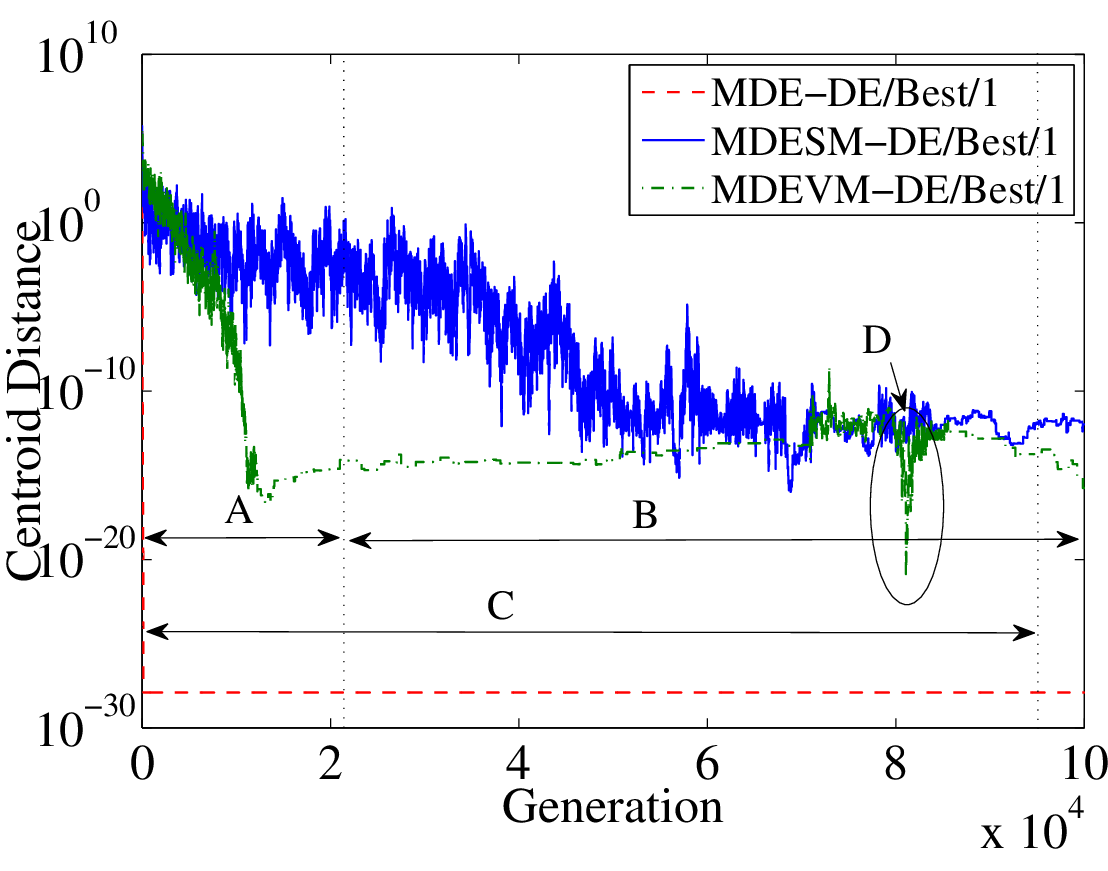}
                \caption{$f_{21}$: DE/Best/1 Scheme}
                \label{fig:}
        \end{subfigure}%
        \begin{subfigure}[b]{0.32\textwidth}
                 \includegraphics[width=1\linewidth]{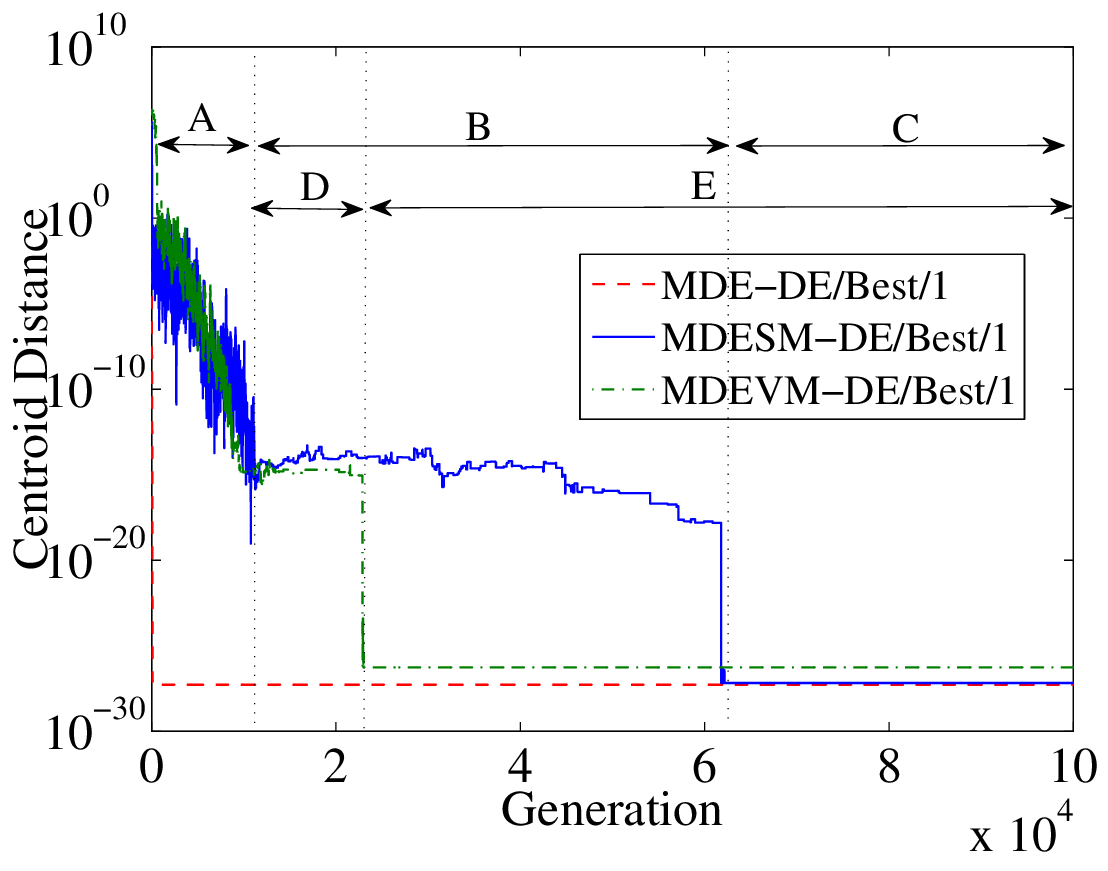}
                \caption{$f_{22}$: DE/Best/1 Scheme}
                \label{HNF20}
        \end{subfigure}%        
        
        \caption{Performance comparison and population centroid-based distance diversity analysis among the MDE, MDESM, and MDEVM schemes for the maximum number of function calls $NFC_{Max}=5000D$, dimension $D=100$, population size $N_{P}=5$, and DE/Rand/1 and DE/Best/1 mutation schemes.}\label{diversity_anal} 
\end{figure*}

The VRMF method can empower the MDE algorithm to escape trapping in local optima and decrease the stagnation risk. In order to analyze the effect of randomization of mutation factor on the population diversity by considering the centroid diversity measure and performance of the MDE algorithm, the best-value-so-far and population diversity plots of the MDE, MDESM, and MDEVM methods are presented for composite functions $f_{20}$ to $f_{22}$ in Figure \ref{diversity_anal}. The simulations are conducted for dimension $D=100$, population size $N_{P}=5$, and schemes  ``DE/Rand/1" and ``DE/Best/1". Conductive to have a better sense of analysis, the maximum number of function calls is considered $NFC_{Max}=5000D$. 

The MDEVM method for the mutation scheme ``DE/Rand/1" has the best performance for the function $f_{20}$ as shown in Figure \ref{diversity_anal}a, denoted by ``B". The population diversities in Figure \ref{diversity_anal}d and for the ``DE/Best/1" mutation scheme in Figure \ref{diversity_anal}j, clearly show that while the MDE and MDESM methods for both mutation schemes are stagnated, due to almost static large value of centroid diversity value, the MDEVM method for the ``DE/Rand/1" has escaped from the stagnation denoted by region ``A" while trying to converge in generations denoted by region ``B". When the diversity is high, and the performance of algorithm in finding the solution is almost static with respect to the best-value-so-far measure, the population is considered stagnated. For situation of trapping in a local minimum, the population is not divert and the diversity is low, while having a poor best-value-so-far performance. 

For the $f_{21}$ case, the MDEVM method using the ``DE/Rand/1" and ``DE/Best/1" schemes has the best performance, as shown in Figure \ref{diversity_anal}b and Figure \ref{diversity_anal}h. The MDE algorithm is trapped in local minimum for both mutation schemes, while the MDESM method has better capability to escape from both stagnation and local optimum trapping, denoted by region ``C" in Figure \ref{diversity_anal}e and Figure \ref{diversity_anal}k. The MDEVM has the best best-value-so-far for both mutation schemes. For the ``DE/Rand/1" mutation scheme, the population's diversity shows a similar convergence trend to the MDESM method, but has achieved a much better best-value-so-far at the beginning generations (i.e., exploration phase) and then trapped in the local minimum, as denoted in region ``C" of Figure \ref{diversity_anal}b. The same performance is obvious for the ``DE/Best/1" mutation scheme as shown in Figure \ref{diversity_anal}h, where in region ``A" it is converged to a solution. The corresponding diversity measure is well-illustrated in Figure \ref{diversity_anal}k. In region ``A", which is the exploration phase, the population's diversity is decreased and it is converged, as shown in region ``B". In ``D", it has trapped but recovered fast to the same level as region ``B". 

The exploring power of VRMF is well illustrated for the benchmark function $f_{22}$ as shown in Figure \ref{diversity_anal}.c and Figure \ref{diversity_anal}.i. In Figure \ref{diversity_anal}.c, it is clear that the VRMF technique has escaped the DE-algorithm from stagnation (denoted by ``A") approximately at $NFC=3\times10^{5}$ and with a sudden movement, as denoted by region ``B", it has reached a better performance than the other methods in region ``C". This is clearly shown in Figure \ref{diversity_anal}f, that the MDEVM algorithm is rescued from stagnation (region ``A") and gradually converging as shown in regions ``B" and ``C". This is while the MDE algorithm is completely trapped in a local minimum, since its best-value-so-far remains constant for all $NFCs$ and the population diversity is extremely low for all generations, i.e. almost $1e-28$ in Figure \ref{diversity_anal}.f. The MDESM has tried to converge (part ``D" of Figure \ref{diversity_anal}f) to the solution as presented in part ``E" of Figure \ref{diversity_anal}.c. However, its exploration is stopped as shown in parts ``E" and ``E" of the Figure \ref{diversity_anal}.c and Figure \ref{diversity_anal}.f, respectively, and no further improvements are achieved. For the ``DE/Best/1" mutation scheme, the MDE is trapped in a local minimum similar to the ``DE/Best/1" mutation scheme, as shown in Figure \ref{diversity_anal}.i and  Figure \ref{diversity_anal}.l. The MDESM has achieved better performance by converging its population toward a solution as denoted by regions ``C" and ``A" in Figure \ref{diversity_anal}i and Figure \ref{diversity_anal}.l, respectively. In further generations, although it has spent some time in generations denoted by region ``B" in Figure \ref{diversity_anal}l to find a better solution, but it has been trapped finally in a local minimum as illustrated in part ``C" of the Figure \ref{diversity_anal}l. The MDEVM has experienced the similar trend as the MDESM (regions  ``A", ``D", and ``E" for centroid diversity in Figure \ref{diversity_anal}l), but with better performance from region ``A" toward region ``B" of Figure \ref{diversity_anal}i.

The centroid-based diversity measure along the best-so-far-value analysis clearly have demonstrated performance of the MDE, MDESM, and MDEVM algorithms in stagnation and local optimum trapping scenarios. The results clearly indicate a successful performance of the VRM approach in delivering diversity into the population. Particularly that after some generations where the algorithm is trapped in local optimum or stagnated, it is rescued and moved toward better solutions, while the other algorithms could not survive.

\section{Conclusion and Future Work}
In evolutionary algorithms (EAs), population size is critical in term of providing diversity into searching procedure. Particularly in the differential evolution (DE), where correct selection of the mutation factor is also a crucial parameter in delivering diversity into the population. Normally larger population sizes provide higher diversity with higher computational cost, which can provide less chance of stagnation and premature convergence due to its high exploration capability. Also, DE can generate a limited number of mutant vectors by using a constant mutation factor. The DE-algorithm with small population size, MDE algorithm, convergence to a solution is faster than standard DE algorithm. Yet, the chance of stagnation and premature convergence increases too. To avoid such situations, diversity should be increased while keeping the convergence speed of algorithm high. The crossover technique is one of the method to inject diversify into the population, where in conjunction with a better mutation scheme it can provide a higher diversity and possible faster finding of solution.

In this paper, we have proposed an enhanced version of the micro-DE (MDE) algorithm based on the important capability of the mutation factor to provide diversity in the population, i.e. the micro-differential evolution using vectorized random mutation factor (MDEVM) algorithm. In this approach, in contrast to the standard MDE, the mutation factor $F$ is selected randomly for each decision variable of each individual in the population. In this case, the population can provide much higher diversity during the search process. In order to analyze the performance of the proposed MDEVM algorithm, we have conducted experiments for different schemes of the mutation factor. The results demonstrate that the proposed MDEVM method is capable of solving complex optimization problems with very small population size and has competitive performance with the $\mu$JADE approach.  

Since the population size of MDEVM is small, the proposed parallel version of MDEVM can be implemented such a way that can evaluate a group of individuals on one central processing unit (CPU). As an example for a population size of four, each individual can conduct processing on a core of a quad-core CPU machine, where most of todays' smart devices are equipped with such processors. In order to design fast but reliable optimization algorithms to tackle with real-time applications, mostly in embedded systems, micro-algorithms can be one of the promising approaches. Particularly, implementation on field-programmable gate arrays (FPGAs) which can provide low-power consumption capabilities for certain applications. It is also interesting to investigate adaptive version of the MDEVM, where the randomness can be under control with different probability distribution through progress of the population. The compact versions of DE have a lot in common with the micro approaches. The same idea is worth to work on using small virtual populations for further research.





\end{document}